\documentclass[twoside,11pt]{article}
\usepackage[abbrvbib, preprint]{jmlr2e}

\usepackage{amsmath} 
\usepackage{mathtools}
\usepackage{algorithm}
\usepackage{algpseudocode}
\usepackage{hyperref}
\usepackage{pifont}
\usepackage{graphicx}
\usepackage{tabularx}
\usepackage{multirow}
\usepackage[caption=false]{subfig}
\usepackage{tikz}
\usepackage{pgfplots}
\usepgfplotslibrary{fillbetween}
\pgfplotsset{compat=newest}
\usepackage{pgfplotstable}
\usepackage{amsmath}
\graphicspath{ {./graphics/} }
\usepackage{booktabs}
\usepackage{xcolor}
\usepackage{mathtools, nccmath}
\usepackage{cleveref}
\usepackage{bbm}
\usepackage{multirow}
\usepackage{ifthen}
\usepackage{wrapfig}

\newboolean{showfigure}

\newcommand{\unc}{u} 
\newcommand{\Unc}{\mathcal{U}} 
\newcommand{\neval}{n_{\mathrm{eval}}}
\newcommand{\nm}{n_{\mathrm{m}}}
\newcommand{\nmi}{n_{\mathrm{m},i}}
\newcommand{\nout}{n_{\mathrm{out}}}
\newcommand{\nunc}{n_{\mathrm{u}}}
\newcommand{\ny}{n_{\mathrm{y}}}
\newcommand{\ntheta}{n_{\mathrm{\theta}}}
\newcommand{\YCP}{\mathcal{Y}_{\mathrm{CP}}}
\newcommand{\YIPM}{\mathcal{Y}_{\mathrm{IPM}}}
\newcommand{\YZCP}{\mathcal{Y}_{\mathrm{ZCP}}}
\newcommand\xSet{1.1}
\newcommand\ySet{0.9}
\newcommand\noutO{*}
\newcommand\noutT{triangle*}

\newtheorem{problem}{Problem}

\crefname{equation}{}{}
\crefname{figure}{Fig.}{Figs.}
\crefname{section}{Sec.}{Secs.}
\crefname{table}{Tab.}{Tabs.}
\crefname{algorithm}{Alg.}{Algs.}
\crefname{definition}{Def.}{Defs.}
\crefname{problem}{Problem}{}
\crefname{subproblem}{Problem}{}
\crefname{theorem}{Thm.}{Thms.}
\crefname{proposition}{Prop.}{Props.}
\crefname{lemma}{Lem.}{Lems.}
\crefname{property}{Property}{}
\crefname{appendix}{Appendix}{}

\definecolor{matlabBlue}{rgb}{0,	0.447000000000000,	0.741000000000000}
\definecolor{matlabRed}{rgb}{0.850000000000000,	0.325000000000000,	0.0980000000000000}
\definecolor{matlabYellow}{rgb}{0.929000000000000,	0.694000000000000,	0.125000000000000}
\definecolor{matlabPurple}{rgb}{0.494000000000000,	0.184000000000000,	0.556000000000000}
\definecolor{matlabGreen}{rgb}{0.466000000000000,	0.674000000000000,	0.188000000000000}

\definecolor{mycolor1}{rgb}{0.85000,0.32500,0.09800}
\definecolor{mycolor2}{rgb}{0.92900,0.69400,0.12500}
\definecolor{mycolor3}{RGB}{25,	107,36}
\definecolor{mycolor4}{RGB}{17,49,79}
\definecolor{mycolor5}{rgb}{0.00000,0.44700,0.74100}
\definecolor{mycolor6}{RGB}{169,142,206}
\definecolor{mycolor7}{rgb}{0.3010,    0.7450,    0.9330}

\DeclareMathOperator*{\argmin}{argmin}

\newcommand{\round}[1]{\ensuremath{\lfloor#1\rceil}}

\DeclareFontFamily{U}{mathx}{\hyphenchar\font45}
\DeclareFontShape{U}{mathx}{m}{n}{
      <5> <6> <7> <8> <9> <10>
      <10.95> <12> <14.4> <17.28> <20.74> <24.88>
      mathx10
      }{}
\DeclareSymbolFont{mathx}{U}{mathx}{m}{n}
\DeclareFontSubstitution{U}{mathx}{m}{n}
\DeclareMathAccent{\widecheck}{0}{mathx}{"71}
\DeclareMathAccent{\wideparen}{0}{mathx}{"75}

\allowdisplaybreaks 

\usepackage{lastpage}
\jmlrheading{26}{2025}{1-\pageref{LastPage}}{5/25; Revised
11/25}{11/25}{25-1161}{Laura Lützow, Michael Eichelbeck, Mykel J. Kochenderfer, Matthias Althoff}
\ShortHeadings{Zono-Conformal Prediction}{Lützow, Eichelbeck, Kochenderfer, and Althoff}

\firstpageno{1}

\begin{document}
\title{Zono-Conformal Prediction: Zonotope-Based Uncertainty Quantification for Regression and Classification Tasks
}

\author{\name Laura Lützow$^{1,2}$
\email laura.luetzow@tum.de
\AND
\name Michael Eichelbeck$^{1}$
\email michael.eichelbeck@tum.de
\AND
\name Mykel J. Kochenderfer$^2$
\email mykel@stanford.edu
\AND
\name Matthias Althoff$^{1}$
\email althoff@tum.de
\AND
\addr $^1$School of Computation, Information and Technology, Technical University of Munich, Germany\\
\hphantom{$~^2$~~}and Munich Center for Machine Learning, Germany\\
$~^2$Department of Aeronautics and Astronautics, Stanford University, USA
}

\editor{Chris Oates}

\maketitle

\begin{abstract}%
Conformal prediction is a popular uncertainty quantification method that augments a base predictor to return sets of predictions with statistically valid coverage guarantees.
However, current methods are often computationally expensive and data-intensive, as they require constructing an uncertainty model before calibration. 
Moreover, existing approaches typically represent the prediction sets with intervals, which limits their ability to capture dependencies in multi-dimensional outputs. 
We address these limitations by introducing zono-conformal prediction, a novel approach inspired by interval predictor models and reachset-conformant identification that constructs prediction zonotopes with assured coverage. 
By placing zonotopic uncertainty sets directly into the model of the base predictor, zono-conformal predictors can be identified via a single, data-efficient linear program.
While we can apply zono-conformal prediction to arbitrary nonlinear base predictors, we focus on feed-forward neural networks in this work. 
Aside from regression tasks, we also construct optimal zono-conformal predictors in classification settings where the output of an uncertain predictor is a set of possible classes.
We provide probabilistic coverage guarantees and present methods for detecting outliers in the identification data. 
In extensive numerical experiments, we show that zono-conformal predictors are less conservative than interval predictor models and standard conformal prediction methods, while achieving a similar coverage over the test data.

\end{abstract}


\setboolean{showfigure}{true}

\section{Introduction}

Prediction models are increasingly deployed in safety-critical domains, such as autonomous vehicles, healthcare, and robotics. 
In these domains, models must not only be accurate and robust but also quantify the uncertainty in their predictions to ensure informed decision-making and safety guarantees. Uncertainty quantification methods address this critical need by providing measures of confidence in the outputs of a model, enabling systems to operate more safely under uncertainty.

In this work, we propose a novel set-based method for uncertainty quantification in multi-output prediction models. Our method applies to highly nonlinear models, such as neural networks, and can be used for both regression and classification tasks. Importantly, our approach does not require assumptions about the data distribution while ensuring that the returned prediction sets cover the true output with a user-defined probability.

Traditionally, this problem has been addressed using conformal prediction techniques \citep{vovk2015crossconformalpredictors}. However, most conformal prediction approaches require two disjoint data sets: one for identifying an uncertainty model and one for establishing coverage guarantees via calibration. This approach often leads to high computational complexity and requires large amounts of data. In contrast, our method integrates uncertainty modeling and calibration into a single optimization problem using only one data set.
To achieve this, we place uncertainties directly into the model of the base predictor and calibrate their sizes using a linear program. This draws inspiration from the interval predictor model framework \citep{campi2009intervalPred}, where interval parameters are computed to enclose all data points while minimizing the width of the prediction sets. 

Another limitation of many conformal prediction techniques and interval predictor models is their reliance on intervals as prediction sets. While computationally efficient, intervals lack flexibility in representing complex shapes and cannot capture dependencies between multiple outputs. To address this limitation, we employ zonotopes as our set representation. Zonotopes, a class of centrally symmetric polytopes, offer a favorable trade-off between expressiveness and computational efficiency, making them popular in set-based computations for control, verification, and identification tasks. Although we focus on zonotopes, the proposed framework can also accommodate other set representations.
Our approach builds on recent developments in reachset-conformant identification \citep{luetzow2025reachset}, where zonotopic uncertainty sets are identified to ensure that the reachable set of the model contains all system measurements. We extend this framework beyond its traditional application in dynamical systems identification to encompass general prediction models in both regression and classification settings, while providing robust coverage guarantees.

\subsection{Related Work} \label{subsec:related}

A broad spectrum of research has addressed uncertainty quantification in machine learning \citep{kabir2018neural,abdar2021review,cartagena2021review}. Probabilistic approaches dominate the field, which typically involve modeling the conditional probability distribution of the outputs using Bayesian methods \citep{goan2020bayesian} or ensemble-based techniques \citep{khosravi2015constructing,lakshminarayanan2017ensembles,gal2016dropout}.
While powerful, these methods often rely on strong assumptions, involve substantial computational costs, and may lack formal coverage guarantees.
This work focuses instead on set-based uncertainty quantification methods, which provide prediction sets that contain the true output with a given probability. These approaches typically require fewer assumptions and can offer formal, distribution-free guarantees.
One of the most prominent frameworks in this category is conformal prediction, which we review subsequently. 
We also discuss interval predictor models and reachset-conformant identification, which inspire the proposed zono-conformal prediction approach.

\subsubsection{Conformal Predictors (CPs).}

Conformal prediction is a distribution-free uncertainty quantification method that transforms point-wise predictions into prediction sets with a user-defined coverage \citep{vovk1999randomness,Lei2013distributionfree}.
Its generality and compatibility with any underlying prediction model have made it popular in modern machine learning. 
While conformal prediction is well studied for univariate outputs, extending it to multi-output settings remains challenging.
The straightforward extension of CPs to multi-dimensional outputs corrects the marginal coverage (e.g., using Bonforonni method) and estimates a prediction interval for each response separately \citep{stankeviciute2021conformal}, resulting in hyper-rectangular and often very conservative prediction sets \citep{feldman2023multiQuantile}. 
	More advanced methods attempt to reduce this conservatism, but often require large data sets or structural assumptions on the uncertainty.

	One line of research explicitly models dependencies between outputs. For example, \citet{messoudi2021copula} employ copulas to capture output correlations, and related ideas have been applied to multivariate time-series forecasting \citep{sun2024copula}. 
	Graph-based deep learning methods have also been explored for quantifying the uncertainty in correlated time-series forecasting \citep{cini2025relational}.
	Another direction focuses on more flexible prediction set shapes.
	Ellipsoidal prediction sets can be computed by estimating the covariance matrix from the training data \citep{messoudi2022ellipsoidal,xu2024cpMulti}.
	\citet{gray2025guaranteed} uses zonotopes to construct a set of functions that contains the true function with a user-defined probability. 
	Flexible nonconvex shapes can be obtained by training a conditional variational auto-encoder \citep{feldman2023multiQuantile} or employing conditional normalizing flows \citep{fang2025contra,luo2025volumesortedpredictionsetefficient}.
	Alternative formulations optimize parameterized shape template functions \citep{tumu2024MultiModalCP} and prediction regions defined by norm balls \citep{braun2025minimumvolumeconformalsets}.
	A comprehensive comparison of multi-output CPs is provided by \citet{dheur2025unifiedcomparativestudygeneralized}.

Most existing approaches decouple the design of prediction set shapes from the calibration step, requiring two separate data sets to guarantee coverage.
In this work, we propose a more data-efficient alternative that couples both steps and ensures coverage guarantees using the scenario approach \citep{calafiore2006scenario}. Although conformal prediction and scenario theory have largely developed as separate research directions, recent works \citep{coppola2024scenarioCP,lin2024verificationScenarioCP,osullivan2025bridgingconformalpredictionscenario} highlight their deep connections, showing that conformal prediction guarantees can be recovered within the framework of scenario theory.
Relatedly, \citet{deAngelis2021consonantPredictiveBeliefs} use scenario theory to construct rectangular predictive sets characterizing the distribution of multivariate data.

\subsubsection{Interval Predictor Models (IPMs).}
The scenario approach has also been central in establishing probabilistic guarantees for interval predictor models \citep{campi2009intervalPred}, which equip simple regression models with prediction intervals.
The fundamentals of IPMs lie in interval regression analysis, where interval parameters are optimized through linear programming to ensure that the predicted intervals contain the observed outputs while minimizing interval width \citep{ishibuchi1990interval}. 
For linear-parameter models, spherical or hyper-rectangular parameter sets are estimated with linear programming, while ellipsoidal parameter sets are identified by solving convex optimization problems \citep{campi2009intervalPred}.
For nonlinear models, such as single-layer neural networks, parameters can be estimated through incremental nonlinear programming approaches, as each observation introduces non-convex constraints \citep{campi2015nonconvex}.
Alternatively, one can first map every input-output pair on a collection of parameter points and then search for a parameter set that tightly encloses parameters corresponding to all scenarios \citep{crespo2021interval}.
For arbitrary neural networks, an additive uncertainty interval can be directly computed from the maximum prediction error after optimizing the minimax criterion during training \citep{garatti2019interval,sadeghi2019efficientTraining}.
A survey of IPMs is provided by \citet{rocchetta20211soft}.
Furthermore, a comparison with bounded-error approaches, which compute the set of parameters consistent with a predefined error bound and the observed data, is presented in \citet{blesa2011robustFault}.

\subsubsection{Reachset-conformant Identification.}

Uncertainties in dynamical systems can be quantified through reachset-conformant identification.
Originally introduced by \citet{althoff2012reachability} and formally defined by \citet{roehm2016reachset}, reachset conformance is a relation between a model and another system, which guarantees the transference of safety properties by ensuring that the model can mimic the behavior of the system \cite[Prop.~1]{roehm2019conformance}. 
In reachset-conformant identification, reachset conformance for the observed system behavior is established by injecting nondeterminism---here referred to as uncertainties---into the model such that all measurements of the system are reachable. 
Convex programs can be used to compute zonotopes with minimum size, which enclose the necessary uncertainties for establishing reachset conformance \citep{gruber2023scalable}, while linear programs can identify enclosing zonotopes, which lead to the smallest reachable sets \citep{Liu2023conf,luetzow2024generator}.
The framework has also been extended to nonlinear models using Taylor series expansions \citep{luetzow2025reachset} and scaled to larger data sets using recursive algorithms \citep{luetzow2025recursive}.

\subsection{Contributions}
In this work, we propose zono-conformal predictors, a novel framework that unifies and generalizes ideas from conformal prediction, IPMs, and reachset-conformant identification. 
Our contributions are:
\begin{itemize}
	\item \textbf{Generalized framework:} We introduce zono-conformal predictors, which extend IPMs from one-dimensional prediction intervals to multi-dimensional zonotopic prediction sets. In contrast to IPMs, our framework (i) supports multi-output prediction with explicit modeling of dependencies, (ii) efficiently handles nonlinear base predictors via a principled linearization and uncertainty allocation strategy, and (iii) extends to classification tasks.
	\item \textbf{Probabilistic guarantees:} We propose novel methods for detecting outliers in the data and provide probabilistic guarantees on the reliability of the identified predictors.
	\item \textbf{Practical impact:} Zono-conformal predictors can replace multi-output conformal prediction across a wide range of applications.
	Unlike most existing methods, they require only a small set of calibration data to produce adaptive prediction sets with valid coverage guarantees.
	Extensive numerical experiments demonstrate that zono-conformal predictors achieve less conservative prediction sets than both classical conformal prediction and IPMs. 
\end{itemize}

\Cref{sec:preliminaries} introduces the notation, the necessary background on conformal prediction and IPMs, and zonotopes. 
\Cref{sec:problem} presents the problem statement.
\Cref{sec:zpm} defines zono-conformal predictors and proposes efficient construction techniques for both regression and classification tasks.
\Cref{sec:coverage} outlines three outlier detection strategies and derives probabilistic coverage guarantees for the identified predictors.
\Cref{sec:experiments} compares the conservativeness and coverage of zono-conformal predictors with standard CPs and IPMs in extensive experiments on synthetic and real-world data sets.
Additional details on the data sets are provided in \cref{sec:datasets}, and the results of ablation studies are discussed in \cref{sec:ablationStudies}.


\section{Preliminaries} \label{sec:preliminaries}
This section lays the groundwork for the development of zono-conformal predictors by first introducing the notation used throughout the paper. We then recall two key set-based uncertainty quantification frameworks---conformal prediction and IPMs---which form the conceptual basis for our approach, and explain how coverage guarantees can be derived using scenario theory. Finally, we introduce zonotopes as a flexible and computationally efficient set representation. 
While reachset-conformant identification is not detailed in this section due to its distinct application domain, its methodological influence on our approach will be highlighted and discussed in subsequent sections.

\subsection{Notation}
We denote matrices by upper-case letters, vectors and scalars by lower-case letters, and sets by calligraphic letters.  
Furthermore, we use $\mathbf{1}$, $\mathbf{I}$, and $\mathbf{0}$, respectively, for a vector filled with ones, the identity matrix, and a matrix of zeros with proper dimensions.
The $j$-th element of a vector $b\in \mathbb{R}^{o}$ is denoted by $b_{(j)}$, while $A_{(\cdot,j)}$ and $A_{(i,j)}$ denote the $j$-th column and the $i$-th element of the $j$-th column of the matrix $A\in \mathbb{R}^{o\times n}$, respectively.
Taking the absolute value of each element of the matrix $A$ is denoted by $|A|\in \mathbb{R}^{o\times n}$.
The vertical concatenation of the vectors or matrices $A_i\in \mathbb{R}^{o_i\times n}$, $i\in\mathcal{I}\subset \mathcal{N}$, is denoted by $\mathrm{vert}_{i\in\mathcal{I}}(A_i)$.
The maximum over all elements of the vector $b\in \mathbb{R}^{o}$ is denoted by $\max_{i} (b_{(i)})\in \mathbb{R}$. 
Additionally, we use the notation $\mathrm{diag}(b)$ for a diagonal matrix with the elements of the vector $b$ on its main diagonal.
The operation $\round{x}$ rounds $x\in\mathbb{R}$ to the nearest integer and $\lfloor \cdot \rfloor$ and $\lceil \cdot \rceil$ denote the floor and the ceil function, respectively. 
The function $\mathbbm{1} \{\text{expression}\}$ returns the value 1 if the boolean expression is true and 0 otherwise.
The Minkowski sum of the two sets $\mathcal{X}_1,\mathcal{X}_2 \subset \mathbb{R}^{n}$ is defined as $\mathcal{X}_1 \oplus \mathcal{X}_2 = \{x_1 + x_2 \mid x_1 \in \mathcal{X}_1,~x_2\in \mathcal{X}_2\}$, while the linear transformation of $\mathcal{X}_1$ using the matrix $A \in \mathbb{R}^{o \times n}$ is defined as $A\mathcal{X}_1 = \{Ax \mid x \in \mathcal{X}_1\}$. The set difference of $\mathcal{X}_1$ and $\mathcal{X}_2$ is given by $\mathcal{X}_1\setminus\mathcal{X}_2=\{x  \mid x \in \mathcal{X}_1 \text{ and } x \notin \mathcal{X}_2\}$.

	\subsection{Set-Based Uncertainty Quantification}\label{subsec:coverageUQ}
	We consider an unknown data-generating process producing samples $(x \in \mathbb{R}^{n_{\mathrm{x}}},\, y \in \mathbb{R}^{\ny})$, without making any assumptions about the underlying probability distribution. 
	Given a calibration data set $\mathcal{M}_{\mathrm{cal}}=\{(x^{(m)},y^{(m)})\}_{m=1}^{\nm}$, both IPMs and CPs generate prediction sets $\mathcal{Y}(\cdot)\subseteq\mathbb{R}^{\ny}$ such
	that $\mathcal{Y}(x)$ contains the output $y$ of a new data point $(x,y)$ with a certain probability. We refer to this probability as the coverage $\eta(\mathcal{M}_{\mathrm{cal}})=\mathbb{P}\{y\in \mathcal{Y}(x)\}$. 
	
	\subsubsection{Interval Predictor Models.}
	
	IPMs generate prediction intervals for one-dimensional regression tasks \citep{campi2009intervalPred}. The predictor is typically constructed as
\begin{align*}
	\YIPM(x) = \{\unc^\top \psi(x) \mid \unc \in \Unc\}\subset \mathbb{R},
\end{align*}
where $\unc\in \Unc \subset \mathbb{R}^{n_{\unc}}$ are uncertain coefficients and $\psi(\cdot)\in\mathbb{R}^{n_{\unc}}$ are fixed basis functions.
The uncertainty set $\Unc$ is parameterized by parameters $\theta$, which are identified such that the calibration points $(x,y)\in\mathcal{M}_{\mathrm{cal}} \setminus \mathcal{M}_{\mathrm{out}}$ satisfy $y\in\YIPM(x)$, while minimizing the width of the prediction intervals. 
The set $\mathcal{M}_{\mathrm{out}}\subset\mathcal{M}_{\mathrm{cal}}$ contains potential outliers, which can be identified using a variety of strategies \citep{huang1998robust,campi2009intervalPred}.

A common parametrization uses $\psi(x) = [1~\tilde{\psi}(x)^\top]^\top$ with $\tilde{\psi}(x)\in \mathbb{R}^{\nunc-1}$, where $u_{(1)}$ lies in the interval $[-\gamma,\gamma]$ and the remaining coefficients $u_{(i)}$, $i=2,...,\nunc$ lie in a ball of radius $r$ centered at $c$: 
\begin{align*}
	\Unc=[-\gamma,\gamma]\times \{\tilde{u}\in \mathbb{R}^{\nunc-1} \mid \| \tilde{u}-c\|_2 \leq r\}.
\end{align*} 
The parameter vector $\theta = [\gamma~r ~c^\top]^\top$ is then optimized with the linear program \citep[Problem 1]{campi2009intervalPred}
\begin{subequations}\label{eq:IPMprogram}
\begin{align}
	\argmin_{\theta}~& \mu r + \gamma   \\
    \text{s.t.}~\forall (x,y) \in \mathcal{M}_{\mathrm{cal}}\setminus \mathcal{M}_{\mathrm{out}}\colon~~&y - c^\top\tilde{\psi}(x)-(r\|\tilde{\psi}(x)\|_2+\gamma)  \leq 0,  \\
	-&y + c^\top\tilde{\psi}(x)-(r\|\tilde{\psi}(x)\|_2+\gamma)  \leq 0, 
\end{align}
\end{subequations}
where $\mu$ is a positive weight.
Coverage guarantees for IPMs can be derived via scenario theory (see \cref{subsec:scenario}).

\subsubsection{Conformal Predictors.}

In this work, we focus on split conformal prediction, the most widely used variant of conformal prediction~\citep{angelopoulos2023CPgentleintroduction}.
Conformal prediction provides a framework for constructing prediction sets that achieve the marginal coverage
\begin{align} \label{eq:expectationCP}
    \mathbb{E}\{\eta(\mathcal{M}_{\mathrm{cal}})\}\geq 1-\epsilon.
\end{align}
To build such sets, one first defines a score function $s(x, y)\colon \mathbb{R}^{n_{\mathrm{x}}}\times\mathbb{R}^{\ny} \to \mathbb{R}$, which quantifies how poorly a data point $(x,y)$ can be explained by a point predictor $f\colon \mathbb{R}^{n_{\mathrm{x}}}\to\mathbb{R}^{\ny}$.
Given the calibration scores $\{s(x^{(m)},y^{(m)})\}_{m=1}^{\nm}$, the prediction set for a new input $x$ is constructed as
\begin{align}
	\YCP(x)=\left\{y \mid s(x,y) \leq q_{\epsilon}\right\},
\end{align} 
where $q_{\epsilon}$ is the $1-n_{\mathrm{out}}/\nm$ empirical quantile of the calibration scores and 
\begin{align} \label{eq:nout}
    n_{\mathrm{out}} = \nm-\lceil(\nm +1)(1-\epsilon) \rceil
\end{align} 
is the number of data points violating $s(x,y) \leq q_{\epsilon}$.

The choice of score function depends on the task:
For regression tasks, a common score function is the $p$-norm of the prediction error, i.e., $s(x,y)=\|f(x)-y\|_{p}$  \citep{braun2025minimumvolumeconformalsets}.
The prediction set for this score function becomes the $p$-norm ball around $f(x)$ with radius $q_{\epsilon}$. 
Alternatively, we can construct and calibrate separate models for each output dimension, e.g., using the score function
\begin{align} \label{eq:cp_add_score}
s_{(j)}(x,y)=|f_{(j)}(x)-y_{(j)}|
\end{align} 
for each output dimension $j$, resulting in rectangular prediction sets. However, to obtain a valid joint coverage of $1-\epsilon$, each model must have a marginal coverage of $1-\frac{\epsilon}{\ny}$ \citep{stankeviciute2021conformal}.
More sophisticated score functions can yield more flexible prediction set shapes, but typically require an additional identification data set.

For classification tasks, a popular score function is
\begin{align}\label{eq:cp_addClass_score}
s(x,y)=1-f_{(y)}(x),
\end{align} 
which computes 1 minus the softmax output $f_{(y)}(x)$ of the true class $y$ \citep{angelopoulos2023CPgentleintroduction}.

\subsection{Scenario Approach}\label{subsec:scenario}

The scenario approach \citep{calafiore2006scenario} provides a rigorous probabilistic framework for optimization under uncertainty. It replaces uncertain constraints with a finite set of sampled scenarios and gives explicit, non-asymptotic guarantees on constraint satisfaction. 
These guarantees align conceptually with the Probably Approximately Correct (PAC) learning bounds from statistical learning theory, as both provide finite-sample generalization guarantees that depend on model complexity and the number of available samples \citep{rocchetta2024survey}.

Let $\mathcal{M}_{\mathrm{cal}}=\{\delta^{(1)},\dots,\delta^{(\nm)}\}\subset \Delta^{\nm}$ denote $\nm$ independent samples from an unknown, time-invariant distribution.
We further allow a subset $\mathcal{M}_{\mathrm{out}}\subseteq \mathcal{M}_{\mathrm{cal}}$ of $\nout$ samples to be removed, e.g., to account for outliers. A convex scenario program is 
\begin{subequations}\label{eq:scenarioProgram}
\begin{align}
\argmin_{\theta}~& J(\theta)  \\
\text{s.t.}~\forall\delta \in \mathcal{M}_{\mathrm{cal}}\setminus \mathcal{M}_{\mathrm{out}}\colon~&g(\delta,\theta) \leq 0, 
\end{align}    
\end{subequations}
where $J$ and $g$ are convex in the parameters $\theta \in \mathbb{R}^{\ntheta}$. 
We assume that \cref{eq:scenarioProgram} is non-degenerate and admits a unique optimal solution \citep[Assumptions~1-2]{rocchetta20211soft}.
Since the program depends on the sampled data set $\mathcal{M}_{\mathrm{cal}}$, the optimal solution $\theta^{*}(\mathcal{M}_{\mathrm{cal}})$ is a random variable. 
We quantify its reliability through the random variable $\eta\colon \Delta^{n_{\mathrm{m}}}\to [0,1]$ defined by
\begin{align*}
\eta(\mathcal{M}_{\mathrm{cal}})=\mathbb{P}\{g(\delta, \theta^{*}(\mathcal{M}_{\mathrm{cal}})) \leq 0\},
\end{align*}
which represents the probability that a new sample $\delta$ satisfies the constraints of \cref{eq:scenarioProgram} for the solution $\theta^{*}(\mathcal{M}_{\mathrm{cal}})$ and is referred to as coverage probability in the uncertainty quantification setting. 
A fundamental result of scenario theory guarantees that the target coverage $1-\epsilon \in (0,1)$ is achieved with a probability of at least \citep[Thm.~3]{rocchetta20211soft}
\begin{align}\label{eq:eta}
\mathbb{P}\{\eta(\mathcal{M}_{\mathrm{cal}}) \geq 1-\epsilon \} \geq 1 - \binom{\nout+\ntheta-1}{\nout} \sum_{i=0}^{\nout+\ntheta-1} \binom{\nm}{i}\epsilon^i(1-\epsilon)^{\nm-i}.
\end{align}
The expected value of $\eta$ can be bounded by \citep{osullivan2025bridgingconformalpredictionscenario}
\begin{align}\label{eq:expectedEta}
\mathbb{E}\{\eta(\mathcal{M}_{\mathrm{cal}})\}\geq 1-\binom{\nout+\ntheta-1}{\nout} \frac{\nout+\ntheta}{\nm+1}.
\end{align}

Scenario theory thus provides a natural framework for analyzing the reliability of prediction models. Specifically, the linear program for constructing IPMs in \cref{eq:IPMprogram} can be interpreted as a convex scenario program with $\delta=(x,y)$. Similarly, CP construction can also be formulated as a scenario program with optimization variable $\theta=q_{\epsilon}$, cost function $J(q_{\epsilon})=q_{\epsilon}$, and constraint function $g(x,y,q_{\epsilon})=s(x,y)-q_{\epsilon}$.
Consequently, the formal coverage guarantees in \cref{eq:eta,eq:expectedEta} hold for prediction sets generated by both IPMs and CPs, without any assumptions on the underlying data distribution.
Moreover, \cref{eq:expectedEta} reduces to \cref{eq:expectationCP} whenever $\epsilon \geq \binom{\nout+\ntheta-1}{\nout} \frac{\nout+\ntheta}{\nm+1}$, which holds for conformal predictors with $\ntheta = 1$ and $\nout$ defined as in \cref{eq:nout}.

\subsection{Zonotopes}

In contrast to IPMs and most CPs, we represent the prediction sets using zonotopes, which are 
centrally symmetric convex polytopes. 
\begin{definition}[Zonotopes \citep{kuehn1998wrapping}]\label{def:zonotopes}
A zonotope $\mathcal{Z}\subset \mathbb{R}^n$ can be described by a center vector $c \in \mathbb{R}^{n}$ and a generator matrix $G\in \mathbb{R}^{n \times {\nu}}$:
\begin{align*}
	\mathcal{Z} = \left\{c+\sum_{i=1}^{\nu} \lambda_i G_{(\cdot,i)} \,\middle\vert\,\lambda_i\in[-1,1]\right\}=\langle  c,G \rangle.
\end{align*}
\end{definition}
Since zonotopes are closed under the set operations linear transformation and Minkowski sum and omit a compact representation size, they are used for efficient set-based computations in control \citep{schaefer2024scalable}, reachability analysis \citep{girard2005zonotopes,althoff2007reachability}, state estimation \citep{alamo2003guaranteed,combastel2015zonotopes}, or identification \citep{chabane2014improved}.
The size of a zonotope can be characterized using the interval norm:
\begin{definition}[Interval Norm \citep{althoff2023conf}]\label{def:intNorm}
The interval norm for the zonotope $\mathcal{Z}=\langle c,G\rangle$ is defined as the absolute sum over all elements of $G$:
\begin{align*} 
	\|\mathcal{Z}\|_I  = \mathbf{1}^\top |G| \mathbf{1}.
\end{align*}

\end{definition}

\section{Problem Statement}\label{sec:problem}

We introduce zono-conformal predictors (ZCPs), which 
return multi-dimensional zonotopic prediction sets $\YZCP(x)\subseteq \mathbb{R}^{\ny}$ that contain the true output with a certain probability.
The model $\YZCP(\cdot)$ is built from the data $\mathcal{M}_{\mathrm{cal}}=\{(x^{(m)},y^{(m)})\}_{m=1}^{\nm}$ by ensuring each prediction set $\YZCP(x^{(m)})$ contains the corresponding output $y^{(m)}$. At the same time, we minimize a proxy for the volume of the prediction sets using the function $\mathrm{size}(\cdot)$, evaluated for inputs $\{x_{\mathrm{eval}}^{(m)}\}_{m=1}^{\neval}$, which are representative samples from the set of admissible inputs. 
Minimizing a proxy for the volume of the prediction sets ensures that the predictions are not too conservative and, thus, informative.
For standard regression tasks, where the observed outputs $y^{(m)}$ correspond to continuous values, the problem can be formalized as follows:

\vspace{0.2cm}
\begin{problem}[Uncertainty Quantification for Regression]\label{prob:regression}\sloppy
Given a regression task with data $\mathcal{M}_{\mathrm{cal}}=\{(x^{(m)},y^{(m)})\}_{m=1}^{\nm}$, we want to find a model $\YZCP(\cdot)$ that solves
\begin{subequations}
	\begin{align}
		\min_{\YZCP(\cdot)} ~&\sum_{m=1}^{\neval} \mathrm{size}(\YZCP(x_{\mathrm{eval}}^{(m)})) \label{eq:regr_cost}\\
		\mathrm{s.t.}~\forall  m\in\{1,\dots,\nm\}\colon ~ &y^{(m)} \in \YZCP(x^{(m)}). \label{eq:regr_constr}
	\end{align}
\end{subequations} 
\end{problem}

Furthermore, we propose to use ZCPs in classification tasks. Here, the observed output $y^{(m)}\in\{0,1\}^{\ny}$ is a binary vector, whose $i$-th element has the value 1 if $x^{(m)}$ can be classified to the $i$-th class and 0 otherwise.
We assume that the output of the available point predictor $f(x)\in \mathbb{R}^{\ny}$ is a vector, indicating a classification to the $i$-th class, if the $i$-th element of $f(x)$ is bigger than or equal to all other elements. 
A non-deterministic classifier, on the other hand, returns a prediction set $\YZCP \subset \mathbb{R}^{\ny}$, which can encode multiple classes. 
By introducing the functions 
\begin{align}\label{eq:classes}
\mathrm{classes}(\mathcal{Y})=\bigl\{i \mid \exists y \in \mathcal{Y}\colon y_{(i)}=\max_j y_{(j)}\bigr\} \text{  and  }    
\mathrm{classes}(y)=\bigl\{i \mid y_{(i)}=\max_j y_{(j)}\bigr\},
\end{align}
which map a set $\mathcal{Y}$ or a vector $y$ onto the encoded classes, 
the uncertainty quantification problem for classification tasks can be stated as follows:

\vspace{0.2cm}
\begin{problem}[Uncertainty Quantification for Classification]\label{prob:classification} \sloppy
Given a classification task with data $\mathcal{M}_{\mathrm{cal}}=\{(x^{(m)},y^{(m)})\}_{m=1}^{\nm}$, we want to find a model $\YZCP(\cdot)$ that solves
\begin{subequations}
	\begin{align}
		\min_{\YZCP(\cdot)}~& \sum_{m=1}^{\neval} \mathrm{size}(\YZCP(x_{\mathrm{eval}}^{(m)}))& \label{eq:class_cost}\\
		\mathrm{s.t.}~\forall  m\in\{1,\dots,\nm\}\colon~ &\mathrm{classes}(y^{(m)})\subseteq \mathrm{classes}\left(\YZCP(x^{(m)})\right). \label{eq:class_constr}
	\end{align}
\end{subequations}
\end{problem}
\vspace{0.2cm}
Since \cref{prob:regression} and \cref{prob:classification} are complex nonlinear programming problems, we will introduce some additional assumptions on the construction of $\YZCP(\cdot)$ in the following section so that both problems can be solved with linear programming.

Additionally, we want to evaluate the coverage probability of ZCPs. To cater to cases where less conservative prediction sets are required while lower coverage is sufficient, we also describe outlier detection methods that discard anomalous data points.


\section{Zono-Conformal Prediction} \label{sec:zpm}

In this section, we discuss how we can efficiently construct zono-conformal predictors $\YZCP(\cdot)$ from observed data. 
We propose the following procedure:

\begin{enumerate}
\item \textbf{Deterministic Model}: We start with a deterministic prediction model, whose output $f(x) \in \mathbb{R}^{\ny}$ can be described by a nonlinear function of the input $x\in\mathbb{R}^{n_{\mathrm{x}}}$.    
\item \textbf{Uncertainty Placement}: 
We create an augmented function $\tilde{f}(x,\unc)\in \mathbb{R}^{\ny}$, with $\tilde{f}(x,\mathbf{0})=f(x)$, by inserting variables $\unc\in \mathbb{R}^{\nunc}$ into the function $f(x)$ to model the uncertainty of the predictor.
\item \textbf{Uncertainty Quantification}: We identify a zonotopic uncertainty set $\Unc$ from observed data such that the nondeterministic model
\begin{align} \label{eq:Z}
	\YZCP(x)=\{f(x) + \bar{D}(x)\unc \mid \unc \in \Unc\} \subset \mathbb{R}^{\ny},
\end{align}
with $\bar{D}(x)=\nabla_{\unc} \tilde{f}(x,\unc)|_{\unc=\mathbf{0}}$,
solves \cref{prob:regression} in the regression setting or \cref{prob:classification} in the classification setting.
\end{enumerate}
These steps are explained in detail in the following subsections.

\subsection{Deterministic Model}

The deterministic prediction model $f(x)$ can be constructed from prior knowledge about the data generation process or identified from additional data.
In this work, we assume $f(x)$ is already given.
While the uncertainty placement and quantification steps, which are described in the following subsections, can be applied to arbitrary nonlinear models, we will focus on neural network models in the experiments in \cref{sec:experiments}.

\subsection{Uncertainty Placement} \label{subsec:uncertaintyPlacement}

In the zono-conformal prediction framework, we aim to construct prediction sets that adapt to the input rather than remaining constant in size and shape. To achieve this,
we insert variables $\unc\in \mathbb{R}^{\nunc}$ into the deterministic function $f(x)$ to obtain the augmented function $\tilde{f}(x,\unc)$. 
This augmented function retains the original behavior of the deterministic predictor when $\unc=\mathbf{0}$, i.e., $\tilde{f}(x,\mathbf{0})=f(x)$.
The variables $\unc$, referred to as uncertainties, represent components of the predictor whose effect on the output is uncertain.

Uncertainties can be incorporated at different locations within the predictor. One straightforward approach is to directly influence the output by adding output uncertainties $\unc_{\mathrm{y}}\in\mathbb{R}^{\ny}$.
Furthermore, parametric uncertainties $\unc_{\mathrm{p}}\in\mathbb{R}^{n_{\mathrm{p}}}$ can be added in form of uncertain model parameters. 
\begin{example}
Consider the deterministic predictor $f(x)\colon \mathbb{R}^2\to \mathbb{R}^2$,
\begin{align*}
	f(x) = \begin{bmatrix}
		3 x_1 + x_2 \\
		2 x_2^3 
	\end{bmatrix}.
\end{align*}
Inserting output uncertainties $\unc_{\mathrm{y}}$ and parametric uncertainties $\unc_{\mathrm{p}}$ leads, for example, to the augmented function
\begin{align*}
	\tilde{f}(x,\unc) = \begin{bmatrix}
		(3+\unc_{p,1}) x_1 + (1+\unc_{p,3})x_2 \\
		(2+\unc_{p,2}) \bigl((1+\unc_{p,4})x_2\bigr)^3 
	\end{bmatrix} + \unc_{\mathrm{y}},
\end{align*}
with $\unc=[\unc_{\mathrm{p}}^\top~\unc_{\mathrm{y}}^\top]^\top$.
\end{example}
Using only output uncertainties corresponds to constant prediction sets, while introducing parametric uncertainties enables adaptive, input-dependent prediction sets.
For physics-based or interpretable models, output uncertainties are appropriate when the output is always noisy or uncertain (e.g., due to measurement noise), whereas parametric uncertainties should be added for components subject to unknown disturbances or partially known parameters (e.g., friction coefficients depending on an unobserved environment).

However, for complex models, such as neural networks, the optimal placement of uncertainties is generally unknown.
In this case, one must select a subset $\unc\in\mathbb{R}^{n_{\unc}}$ from a large pool of candidate uncertainties $\tilde{\unc}=[\unc_{\mathrm{p}}^\top~\unc_{\mathrm{y}}^\top]^\top\in\mathbb{R}^{n_{\mathrm{p}}+\ny}$, with $\nunc<n_{\mathrm{p}}+\ny$.
Identifying too many uncertainties can lead to increased computational cost and reduced coverage due to overfitting. Conversely, a poorly chosen uncertainty placement strategy can lead to excessive conservatism and, in the worst case, may render the constraints in \cref{eq:regr_constr} or \cref{eq:class_constr} unsatisfiable.
To address this, we propose a simple yet effective strategy:
\begin{enumerate}
\item Select all output uncertainties $\unc_{\mathrm{y}}$ to ensure adequate control over each output dimension.
\item Randomly sample the remaining $\nunc-\ny$ uncertainties from the pool of parametric uncertainties $\unc_{\mathrm{p}}$.
\end{enumerate}
As shown in \cref{subsec:ablationUP}, this strategy consistently outperforms purely random selection and a more complex, deterministic approach using QR-factorization.

\subsection{Uncertainty Quantification} \label{subsec:uncertaintyQuantification}

We construct the zono-conformal predictor as in \cref{eq:Z} by evaluating the first-order Taylor-series approximation of $\tilde{f}(x,\unc)$ at $\unc=\mathbf{0}$ over the uncertainties $\unc\in \Unc$. 
This linearization is motivated by reachset-conformant identification, where the reachable set of a nonlinear system under bounded uncertainty $u\in\mathcal{U}$ is approximated using a Taylor series \citep{althoff2013nonlin}. 
The uncertainty set $\Unc$ is then identified by solving \cref{prob:regression} or \cref{prob:classification}, both of which reduce to linear programs under the following assumptions: 

\begin{itemize}
\item To motivate small and informative prediction sets, we minimize the following cost function, which computes the summed interval norms of randomly rotated versions of the prediction set $\YZCP(x)$:
\begin{align} \label{eq:costRotInterval}
	\mathrm{size}(\YZCP(x)) = \sum_{i=0}^{n_{\mathrm{r}}} \|R_i\YZCP(x)\|_I,
\end{align}
where $n_{\mathrm{r}}\geq 1$ is a user-defined integer, $R_i$, with $i\geq1$, are random orthogonal matrices, and $R_0=\mathbf{I}$.
The standard approach in reachset-conformant identification uses only the unrotated interval norm, i.e., \cref{eq:costRotInterval} with $n_{\mathrm{r}}=0$, but our experiments in \cref{subsec:costRegression} show that incorporating random rotations significantly reduces the conservativeness of the prediction sets.
\item We represent the uncertainty set $\mathcal{U}$ by the zonotope
\begin{align} \label{eq:Unc}
	\Unc = \langle \mathbf{0}, G_{\mathrm{\unc}} \mathrm{diag}(\alpha) \rangle,\quad \alpha \in \mathbb{R}^{\ntheta}_{\geq 0},
\end{align}
where the scaling factors $\alpha$ are the optimization variables, while the unscaled generator matrix $G_{\mathrm{\unc}}\in \mathbb{R}^{\nunc\times\ntheta}$ must be specified by the user. 
As shown by the experiments in \cref{subsec:ablationUP}, defining $G_{\mathrm{\unc}}$ as the identity matrix usually provides good results.
\end{itemize}
Before deriving the linear program formulations for regression and classification, we present a few results needed for both cases.
\begin{lemma}\label{lem:predZonotope}
The prediction set of the zono-conformal predictor in \cref{eq:Z} with the uncertainty set in \cref{eq:Unc} can be described by the zonotope
\begin{align} \label{eq:Z_zonotope}
	\YZCP(x) = \langle f(x), ~\bar{D}(x) G_{\mathrm{\unc}}  \mathrm{diag}(\alpha)\rangle.
\end{align}
\begin{proof}
	Inserting \cref{eq:Unc} into \cref{eq:Z}, results in \cref{eq:Z_zonotope}.      
\end{proof}
\end{lemma}
\begin{lemma}\label{lem:cost}
The interval norm of the rotated prediction set $R_i \YZCP(x)$ of the zono-conformal predictor in \cref{eq:Z} with the uncertainty set in \cref{eq:Unc} can be computed as
\begin{align}\label{eq:cost}
	\| R_i\YZCP(x) \|_I &= \mathbf{1}^\top | R_i\bar{D}(x) G_{\mathrm{\unc}}|  \alpha,
\end{align}            
\begin{proof} 	
	Using \cref{eq:Z_zonotope} and Def.~\ref{def:intNorm}, we obtain
	\begin{align*}
		\| R_i\YZCP(x) \|_I &= \mathbf{1}^\top | R_i\bar{D}(x) G_{\mathrm{\unc}}  \mathrm{diag}(\alpha) | \mathbf{1},
	\end{align*}
	which is equal to \cref{eq:cost} for $\alpha\geq\mathbf{0}$.
\end{proof}
\end{lemma}
\begin{lemma}
A point $y$ is contained in the prediction set $\YZCP(x)$ of the zono-conformal predictor in \cref{eq:Z} with the uncertainty set in \cref{eq:Unc}, i.e., $y\in \YZCP(x)$, iff
\begin{align}\label{eq:containment}
	\exists \bar{\beta}\colon |\bar{\beta}| \leq \alpha,~ 
	\quad y = f(x)  + \bar{D}(x) G_{\mathrm{\unc}} \bar{\beta}.
\end{align}            
\begin{proof}  
	This result can be derived using Def.~\ref{def:zonotopes} and Lem.~\ref{lem:predZonotope} analogously to \citet[Thm.~2]{luetzow2024generator}.
\end{proof}
\end{lemma}


Thus, we obtain the following formulation for regression problems:
\begin{theorem}[Zono-Conformal Regression]\label{thm:regr_id}
\cref{prob:regression} with the cost function in \cref{eq:costRotInterval} can be solved with the zono-conformal predictor given by \cref{eq:Z} and \cref{eq:Unc}, where the optimal scaling factors $\alpha$ are computed with the following linear program:
\begin{subequations}\label{eq:regr_opt}
	\begin{align}
		\argmin_{\alpha,\beta} \sum_{m=1}^{\neval} &\sum_{i=0}^{n_{\mathrm{r}}}\mathbf{1}^\top 
		|R_i\bar{D}(x_{\mathrm{eval}}^{(m)}) G_{\mathrm{\unc}}| \alpha \label{eq:regr_cost_id}\\
		\mathrm{s.t.}~ \forall m\in\{1,\dots,\nm\}\colon \qquad \mathbf{0} &\leq \alpha, \label{eq:regr_constr_pos} \\
		\mathbf{0} &\leq 
		\alpha + \beta_m, \label{eq:regr_constr_idIneq1}\\
		\mathbf{0} &\leq 
		\alpha - \beta_m, \label{eq:regr_constr_idIneq2}\\
		{y}_{\mathrm{\Delta}}^{(m)} &=  \bar{D}(x^{(m)}) G_{\mathrm{\unc}} \beta_m\label{eq:regr_constr_idEq},
	\end{align}
\end{subequations}
with ${y}_{\mathrm{\Delta}}^{(m)} = y^{(m)} - f(x^{(m)})$ and $\beta=[\beta_0^\top,~\ldots,~\beta_{\nm }^\top]^\top$ is a vector of additional optimization variables to enforce the containment constraints.
\begin{proof}
	Evaluating the interval norm of $\YZCP(x)$ over the inputs $x_{\mathrm{eval}}^{(m)}$ with the constraint \cref{eq:regr_constr_pos}, as in \cref{eq:regr_cost}, leads to the linear cost in \cref{eq:regr_cost_id}. 
	Constraints \cref{eq:regr_constr_idIneq1,eq:regr_constr_idIneq2,eq:regr_constr_idEq} can be obtained by using \cref{eq:containment} for the containment constraints in \cref{eq:regr_constr} and introducing a new optimization variable ${\beta}_m$ for each observation $(x^{(m)},y^{(m)})$.            
\end{proof}
\end{theorem}

For classification problems, we assume that each output $y$ is mapped to a single class using $\mathrm{classes}(y)$. To extend zono-conformal predictors to multi-label classification, the calibration data set is constructed by including data points of the form $(x,\mathbf{e}_{i})$ for each class $i$ that is a possible label for the input $x$, where $\mathbf{e}_i$ denotes the $i$-th unit vector.

\begin{theorem}[Zono-Conformal Classification]\label{thm:class_id1}
\cref{prob:classification} with the cost function in \cref{eq:costRotInterval} can be solved with the zono-conformal predictor given by \cref{eq:Z} and \cref{eq:Unc}, where the optimal scaling factors $\alpha$ are computed with the following linear program:
\begin{subequations}\label{eq:class_opt}
	\begin{align}
		\argmin_{\alpha^, \beta} \sum_{m=1}^{\neval} \sum_{i=0}^{n_{\mathrm{r}}}&\mathbf{1}^\top 
		|R_i\bar{D}(x_{\mathrm{eval}}^{(m)}) G_{\mathrm{\unc}}| \alpha  \label{eq:class_cost_id}\\
		\mathrm{s.t.}~ \forall m\in\{1,\dots,\nm\}\colon \qquad \mathbf{0} &\leq \alpha, \label{eq:class_constr_pos} \\
		\mathbf{0} &\leq \alpha + \beta_{m}, \label{eq:class_constr_idIneq1}\\
		\mathbf{0} &\leq \alpha - \beta_{m}, \label{eq:class_constr_idIneq2}\\
		-T_{m} f(x^{(m)}) &\leq T_{m}\bar{D}(x^{(m)}) G_{\mathrm{\unc}} \beta_{m}, \label{eq:class_constr_idIneq3}
	\end{align}
\end{subequations}
with the new optimization variable $\beta=\mathrm{vert}_{m\in\{1,\dots,\nm\}}(\beta_{m})$ and
\begin{align}\label{eq:Ti}
	T_{m} &= \mathrm{vert}_{n_\mathrm{y}}(\mathbf{e}_{\mathrm{classes}(y^{(m)})}^\top) - \mathbf{I}. 
\end{align}
\begin{proof}  
	Evaluating the interval norm of $\YZCP(x)$ over the inputs $x_{\mathrm{eval}}^{(m)}$ with the constraints \cref{eq:class_constr_pos}, as in \cref{eq:class_cost}, leads to the linear cost in \cref{eq:class_cost_id}.     
	Since we have
	\begin{align} \label{eq:y_01}
		y_{(i)}=\begin{cases}
			1 & \text{if $i=  \mathrm{classes}(y)$},\\
			0 & \text{otherwise},
		\end{cases}
	\end{align}
	the constraint in \cref{eq:class_constr} can be written as
	\begin{align*}
		\mathrm{classes}(y^{(m)})\in \mathrm{classes}\bigl(\YZCP(x^{(m)})\bigr) 
		\overset{\text{\cref{eq:classes}}}{\Leftrightarrow}~~& \exists z \in\YZCP(x^{(m)})\colon \max_j z_{(j)} =z_{(\mathrm{classes}(y^{(m)}))}  \\
		\Leftrightarrow~~& \exists z \in\YZCP(x^{(m)})\colon z \leq \mathrm{vert}_{n_\mathrm{y}}(\mathbf{e}_{\mathrm{classes}(y^{(m)})}^\top)z  \\
		\Leftrightarrow~~& \exists z \in\YZCP(x)\colon \mathbf{0} \leq T_m z  \\
		\overset{\text{\cref{eq:containment}}}{\Leftrightarrow}\,~& \exists \beta_{m}\colon |\beta_{m}| \leq \alpha,~
		\mathbf{0} \leq T_{m}(f(x^{(m)}) + \bar{D}(x^{(m)}) G_{\mathrm{\unc}} \beta_{m}),
	\end{align*}
	resulting in \cref{eq:class_constr_idIneq1,eq:class_constr_idIneq2,eq:class_constr_idIneq3}.       
\end{proof}
\end{theorem}


\section{Coverage}\label{sec:coverage}

In conformal prediction, the trade-off between high coverage and smaller prediction sets can be controlled by selecting how many calibration data points should be contained in the corresponding prediction sets. 
This principle also extends to ZCPs. 
We present three methods for identifying outliers, i.e., data points whose removal most effectively reduces the conservatism of ZCPs.
Furthermore, we describe the resulting coverage guarantees of ZCPs and compare them with those of CPs.

\subsection{Outlier Detection} \label{sec:outlier}

Finding the subset of data points whose removal yields the greatest reduction in optimization cost is a combinatorial problem, and solving it through exhaustive search is computationally infeasible. 
In the following sections, we present three efficient methods for outlier detection, which are evaluated and compared in \cref{subsec:ablationOD}.
All three methods lead to significantly smaller optimization costs compared to a naive outlier detection approach, which is not tailored to the zono-conformal prediction framework. 

\subsubsection{Search over Boundary Points} \label{subsubsec:boundary}

\begin{wrapfigure}{r}{0.33\textwidth}
\vspace{-0.2cm}
\centering
\begin{tikzpicture}
	\fill[mycolor3!10] (0,0) -- (2,0) -- (3,1) -- (2,2) -- (0,2) -- (-1,1) -- cycle;
	\draw[mycolor3, thick] (0,0) -- (2,0) -- (3,1) -- (2,2) -- (0,2) -- (-1,1) -- cycle;
	
	\node[anchor=south,font=\footnotesize, mycolor3, text width=2.5cm, text centered] at (-0.5,2) {Prediction set $\YZCP(x)$};
	
	\draw[line width=0.9pt, black!50] (0.95,1.35) -- (1.05,1.45);
	\draw[line width=0.9pt, black!50] (0.95,1.45) -- (1.05,1.35);
	
	\draw[line width=0.9pt, black!50] (0.65,1.85) -- (0.75,1.95);
	\draw[line width=0.9pt, black!50] (0.65,1.95) -- (0.75,1.85);
	
	\draw[line width=0.9pt, black!50] (0.45,1.55) -- (0.55,1.65);
	\draw[line width=0.9pt, black!50] (0.45,1.65) -- (0.55,1.55);
	\node[anchor=north,font=\footnotesize, black!50, text width=3.6cm, text centered] at (0.9,1.3) {Outputs $y$ of non-boundary points};
	
	\draw[line width=0.9pt, red!50!black] (1.45,1.95) -- (1.55,2.05);
	\draw[line width=0.9pt, red!50!black] (1.45,2.05) -- (1.55,1.95);
	
	\draw[line width=0.9pt, red!50!black] (1.95,1.95) -- (2.05,2.05);
	\draw[line width=0.9pt, red!50!black] (1.95,2.05) -- (2.05,1.95);
	
	\draw[line width=0.9pt, red!50!black] (2.25,1.65) -- (2.35,1.75);
	\draw[line width=0.9pt, red!50!black] (2.25,1.75) -- (2.35,1.65);
	\node[anchor=south,font=\footnotesize, red!50!black, text width=3cm, text centered] at (2,2.05) {Outputs $y$ of boundary points};
\end{tikzpicture}
\vspace{-0.6cm}
\caption{Boundary and non-boundary points.}
\label{fig:boundaryPoints}
\vspace{-0.5cm}
\end{wrapfigure}
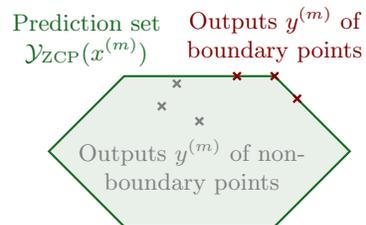

In this work, we define a data point $(x,y)$ as a boundary point if there exists at least one $\alpha_{(j)}$, $j=1,\dots,\ntheta$, such that an infinitesimal reduction in $\alpha_{(j)}$ causes the output $y$ to fall outside its 
corresponding prediction set $\YZCP(x)$, as shown in \cref{fig:boundaryPoints}.
Boundary 
points, therefore, induce constraints in the linear programs \eqref{eq:regr_opt} or \eqref{eq:class_opt} that block any further reduction of the optimization cost at the optimum.
Leveraging this definition, outlier detection can be framed as a search over boundary points, similarly to \cite[Alg.\,1]{campi2009intervalPred}.
Specifically, we construct a search tree rooted at the original problem, where each node represents a modified version in which one data point---a boundary point in the problem of its parent node---has been removed.
This process is repeated for $n_{\text{out}}$ layers, yielding leaf nodes that each exclude $n_{\text{out}}$ data points. The leaf node with the lowest objective value then identifies the optimal set of outliers.

The runtime of this algorithm critically depends on how quickly one can identify boundary points at each node.
By assuming non-degenerate problems \citep[Def.~4]{campi2009intervalPred}, \citet{campi2009intervalPred} identify boundary points as data points whose removal leads to a strict decrease of the optimization cost.
However, this approach does not work for degenerate problems and leads to high computational cost for large data sets, as we have to compute the optimization cost for all possible removals.
Instead, we propose a more scalable method that identifies the set of boundary points for a given node $i$ and the associated calibration set $\mathcal{M}_{\mathrm{cal},i}=\{(x^{(m)},y^{(m)})\}_{m=1}^{\nmi}$ directly from the solution of \cref{eq:regr_opt} for regression tasks or \cref{eq:class_opt} for classification tasks:

\begin{proposition}[Detection of Boundary Points]\label{prop:boundary_id}\sloppy
Boundary points can be identified by computing the vector $\delta$ 
using the following linear program:
\begin{subequations}\label{eq:boundary_opt}
	\begin{align}
		\argmin_{\beta,\delta}~& -\mathbf{1}^\top \delta \label{eq:boundary_cost_id}\\
		\mathrm{s.t.}~ \forall m\in\{1,\dots,\nmi\}\colon ~ &~~~~~0 \leq \delta_{(m)}\\
		&-\alpha^* \leq 
		\beta_m - f \delta_{(m)}, \label{eq:boundary_constr_idIneq1}\\
		&-\alpha^* \leq - \beta_m - f \delta_{(m)}, \label{eq:boundary_constr_idIneq2}\\
		&\cref{eq:regr_constr_idEq} \text{ for regression or }\cref{eq:class_constr_idIneq3} \text{ for classification}, \label{eq:boundary_constr_3}
	\end{align}
\end{subequations}
where $\alpha^*$ is the optimal solution of \cref{eq:regr_opt} or \cref{eq:class_opt}, and the elements of $f\in\mathbb{R}^{\nunc}$ are given by
\begin{align*}
	f_{(j)} = \begin{cases}
		1 & \text{if }\alpha^*_{(j)} > 0,\\
		0 & \text{otherwise}.
	\end{cases}
\end{align*}
The $m$-th data point is a boundary point iff $\delta_{(m)}=0$.
\begin{proof}
	A data point $(x^{(m)},y^{(m)})$ is not a boundary point iff this data point satisfies the containment constraint \cref{eq:regr_constr} for regression tasks or \cref{eq:class_constr} for classification tasks, even when the non-zero elements of the optimal scaling factors $\alpha^*$ are slightly decreased. 
	This occurs if there exists a vector $\beta_m$ satisfying the containment constraints \cref{eq:regr_constr} via constraints \cref{eq:regr_constr_idEq,eq:regr_constr_idIneq1,eq:regr_constr_idIneq2} or \cref{eq:class_constr} via constraints \cref{eq:class_constr_idIneq1,eq:class_constr_idIneq2,eq:class_constr_idIneq3} for $\alpha=\alpha^*$ but whose elements satisfy 
	\begin{align}\label{eq:m_notActive}
		\forall j\colon |\beta_{m(j)}|<\alpha^*_{(j)}~ \text{ or } ~\alpha^*_{(j)}=0.
	\end{align}      
	The linear program \eqref{eq:boundary_opt} encodes this condition by maximizing the slack $\delta_{(m)} = \min_{j\in \mathcal{J}} (\alpha^*_{(j)} - |\beta_{m(j)}|)$, which is required to be greater than or equal to zero, with $\mathcal{J}=\{j \mid \alpha^*_{(j)} > 0\}$. The constraints \cref{eq:boundary_constr_idIneq1,eq:boundary_constr_idIneq2,eq:boundary_constr_3} ensure that each $\beta_m$ remains feasible for the containment constraints under $\alpha = \alpha^*$. A non-zero $\delta_{(m)}$ implies that there exists a vector $\beta_m$ satisfying \cref{eq:m_notActive}, meaning the $m$-th data point is not a boundary point. Conversely, $\delta_{(m)} = 0$ means no such $\beta_m$ exists and the data point is a boundary point.
\end{proof}
\end{proposition}

\subsubsection{Greedy Search over Boundary Points} \label{subsubsec:boundaryGreedy}

To reduce computational costs, a greedy variant of the search strategy described in \cref{subsubsec:boundary} can be employed.
Rather than expanding all nodes at each level, the greedy algorithm only retains the child node with the lowest cost, trading optimality guarantees for significantly improved scalability.

\subsubsection{Detection via Mixed-Integer Linear Program}\label{subsubsec:milp}
Alternatively, we can incorporate the outlier detection into the optimization problems \eqref{eq:regr_opt} and \eqref{eq:class_opt}, resulting in the following mixed-integer linear programs (MILPs):
\begin{proposition}[Zono-Conformal Regression with Outlier Removal]\label{prop:regr_id_OD}\sloppy
If $\nout $ data points are permitted to violate the constraints in \cref{eq:regr_constr}, \cref{prob:regression} can be solved with \cref{thm:regr_id}, where constraint \cref{eq:regr_constr_idEq} is replaced by the constraints 
\begin{subequations}
	\begin{align}
		\mathbf{0} &=  \bar{D}(x^{(m)}) G_{\mathrm{\unc}} \beta_m - {y}_{\mathrm{\Delta}}^{(m)} \rho_{(m)}\label{eq:regr_OD_constr1}, \\
		\nm -\nout & \leq \mathbf{1}^\top \rho, \label{eq:regr_OD_constr2}\\
		\rho &\in \{0,1\}^{\nm }.\label{eq:regr_OD_constr3}
	\end{align}    
\end{subequations}
The vector $\rho\in\mathbb{B}^{\nm}$ is an additional optimization variable, where $\rho_{(m)}$ is equal to 1 if data point $m$ is required to satisfy the constraint in \cref{eq:regr_constr} and 0 if the measurement is considered as an outlier and permitted to violate the constraint in \cref{eq:regr_constr}.  
\begin{proof}  
	The containment $y^{(m)}\in\YZCP(x^{(m)})$ is enforced via \cref{eq:regr_constr_idIneq1,eq:regr_constr_idIneq2,eq:regr_constr_idEq}.
	By multiplying the left-hand side of \cref{eq:regr_constr_idEq} with $\rho_{(m)}$, we obtain \cref{eq:regr_OD_constr1}.
	If $\rho_{(m)}=1$, the new constraint \cref{eq:regr_OD_constr1} is equal to \cref{eq:regr_constr_idEq}, which means the containment $y^{(m)}\in\YZCP(x^{(m)})$ is enforced.
	If $\rho_{(m)}=0$, the constraints \cref{eq:regr_constr_idIneq1,eq:regr_constr_idIneq2,eq:regr_OD_constr1} can be trivially solved with $\beta_m=\mathbf{0}$, which means $y^{(m)}$ is classified as an outlier since the containment $y^{(m)}\in\YZCP(x^{(m)})$ is not enforced. 
	
	The limitation of the values $\rho$ can take is encoded via the integer constraint in \cref{eq:regr_OD_constr3}, while \cref{eq:regr_OD_constr2} ensures that a maximum of $\nout $ measurements are classified as outliers.
\end{proof}
\end{proposition}

\begin{proposition}[Zono-Conformal Classification with Outlier Removal]\label{prop:class_id_OD}\sloppy
If $\nout $ data points are permitted to violate the constraints in \cref{eq:class_constr}, \cref{prob:classification} can be solved with \cref{thm:class_id1}, where constraint \cref{eq:class_constr_idIneq3} is replaced by the constraints 
\begin{subequations}
	\begin{align}
		\mathbf{0} &\leq T_{m}\bar{D}(x^{(m)}) G_{\mathrm{\unc}} \beta_{m} + T_m f(x^{(m)}) \rho_{(m)}, \label{eq:class_OD_constr1} \\
		\nm -\nout & \leq \mathbf{1}^\top \rho, \label{eq:class_OD_constr2} \\
		\rho &\in \{0,1\}^{\nm }. \label{eq:class_OD_constr3}
	\end{align}
\end{subequations}
The vector $\rho\in\mathbb{B}^{\nm}$ is an additional optimization variable, where $\rho_{(m)}$ is equal to 1 if data point $m$ is required to satisfy the constraint in \cref{eq:class_constr} and 0 if the measurement is considered as an outlier and permitted to violate the constraint in \cref{eq:class_constr}.   
\begin{proof}  
	Analogously to the proof of Prop.~\ref{prop:regr_id_OD}.
\end{proof}
\end{proposition}

\subsection{Coverage Guarantees}

The coverage guarantees of IPMs and CPs can be derived from scenario theory \citep{osullivan2025bridgingconformalpredictionscenario}, as discussed in \cref{subsec:scenario}.
Assuming the data points in $\mathcal{M}_{\mathrm{cal}}$ are independently drawn from a time-invariant distribution and independent from the evaluation points $\{x_{\mathrm{eval}}^{(m)}\}_{m=1}^{\neval}$, we can directly apply these results to ZCPs, since \cref{eq:regr_opt,eq:class_opt} are convex scenario programs. Thus, the coverage of ZCPs satisfies \cref{eq:eta,eq:expectedEta}.
Depending on the application, one may either prescribe a desired coverage level and adjust the number of uncertainty parameters $\ntheta$ and outliers $\nout$ accordingly, or fix these parameters in advance and derive the resulting coverage.

Although the guarantees in \cref{eq:eta,eq:expectedEta} formally hold regardless of the output dimension, higher-dimensional outputs typically require more complex models with a larger number of parameters $\ntheta$, which leads to smaller theoretical confidence bounds---see right-hand side of \cref{eq:eta}.
Since the number of identified parameters $\ntheta$ for most CPs is equal to one, while ZCPs often involve a larger $\ntheta$ to capture richer uncertainty structures, we expect higher coverage guarantees for CPs under the same outlier budget $\nout$.
Nevertheless, our experiments in \cref{sec:experiments} demonstrate that ZCPs consistently yield smaller prediction sets while maintaining comparable empirical coverage, suggesting a favorable trade-off between informativeness and coverage.


\section{Experiments} \label{sec:experiments}
We analyze the performance of ZCPs using different synthetic and real-world data sets.
For each task, we train 30 neural networks utilizing tanh activation functions, which will serve as the base predictors $f(x)$.
The network architectures as well as the specifications of the data sets are provided in Tab.~\ref{tab:data}.
More details on the synthetic data sets and the \emph{Photovoltaic} data set are provided in \cref{sec:datasets}.
Before each training, we randomly select 10.000 data points from each data set with more than 10.000 instances.
We randomly split the data into a calibration set, which contains 10\% of the instances, a test set, containing 15\% of the instances, and a training set for training the base predictor, consisting of the remaining 75\% of the instances. 

For the construction of ZPMs, we select $\nunc=\ny+0.1n_{\mathrm{p}}$ uncertainties as described in \cref{subsec:uncertaintyPlacement}, where parametric uncertainties $\unc_{\mathrm{p}}$ are added to all network biases in the hidden layers, we use the cost function from Lem.~\ref{lem:cost} with $n_{\mathrm{r}}=10$, and we detect outliers via greedy search over boundary points as described in \cref{subsubsec:boundaryGreedy}.
Other uncertainty placement strategies, cost functions, and outlier detection methods are evaluated in \cref{sec:ablationStudies}.
Our methods are implemented in the MATLAB toolbox CORA\footnote{The CORA toolbox is available at \href{https://tumcps.github.io/CORA/}{https://tumcps.github.io/CORA/}.}.

\begin{table}[t]
\centering
\begin{tabular}{c c c c c c c}
	\toprule
	\textbf{Task} & $n_{\mathrm{x}}$      & $\ny$  & Network Size  & $n_{\mathrm{p}}$  & $n_{\mathrm{total}}$ & Reference \\ 
	\midrule
	{SD-R1}      & 2           &   2     &  $[64,64]$  &  128  &    $\infty$ & \cref{sec:datasets} \\ 
	{SD-R2}  & 3           &   2      &  $[64,64]$   &  128 &    $\infty$ & \cref{sec:datasets} \\ 
	{Energy}  & 8           &   2      &  $[64,64]$  &  128  &  768 & \cite{tsanas2012energyEfficiency}\\
	{CASP}  & 8           &   2      &  $[64,64]$  &  128  &  45730 & \cite{feldman2023multiQuantile}\\
	{Housing}  & 17           &   2      &  $[64,64]$  &  128  &  21613 & \cite{feldman2023multiQuantile} \\
	{BlogPost}  & 279           &   2      &  $[64,64]$  &  128  &  52397 & \cite{feldman2023multiQuantile} \\
	{SCM20D}  & 61 & 4 & $[64,64]$  &  128 & 8966 & \cite{tsoumakas2011mulan} \\
	{SCM1D}  & 280 & 4 & $[64,64]$  &  128& 9803 & \cite{tsoumakas2011mulan} \\
	{RF1}  & 64           &   8      &  $[64,64]$  &  128  &  9004 & \cite{tsoumakas2011mulan} \\
	{Photovoltaic}  & 49           &   4      &  $[64, 256, 64]$  &  384  &  8639 & \cref{sec:datasets} \\ 
	\midrule
	{SD-C1}  & 2           &   3      &  $[64,64]$  &  128  &    $\infty$ & \cref{sec:datasets} \\
	{SD-C2}  & 3           &   4      &  $[64,64]$  &  128  &    $\infty$ & \cref{sec:datasets} \\
	{Cover type}  & 54           &   7      &  $[128,128]$  &  256  &   581012 & \cite{blackard1999covertypes}\\
	{MNIST}  & 784           &   10      &  $[128,128]$  &  256  &   70000 & \cite{lecun1998mnist}\\
	\bottomrule
\end{tabular}    
\caption{Overview of the different prediction tasks, where the first ten tasks are regression and the last four are classification data sets.} 
\label{tab:data}
\end{table}

\subsection{Regression Tasks}

For regression tasks, we compare ZCPs with the following two baseline methods:
\begin{itemize}
\item IPMs that use the zono-conformal prediction framework to handle nonlinear systems with multi-dimensional outputs while representing the uncertainty sets and prediction sets with multi-dimensional intervals. The resulting predictor is identical to a ZCP that uses a diagonal generator matrix for the uncertainty set $\Unc$ and whose output set is overapproximated by a multi-dimensional interval.
\item CPs that use the prediction error as score function---see \cref{eq:cp_add_score}---to identify a prediction interval for each dimension separately. 
\end{itemize}
Since the more sophisticated conformal prediction approaches, as reviewed in \cref{subsec:related}, require additional data or access to the training data to estimate the shape of the uncertainty sets \citep{cini2025relational,feldman2023multiQuantile,messoudi2022ellipsoidal,xu2024cpMulti,gray2025guaranteed,fang2025contra,luo2025volumesortedpredictionsetefficient},
we do not include them in our comparison.

We identify for each neural network a ZCP, an IPM, and a CP using the calibration data and considering $n_{\mathrm{out}}=0,...,5$ outliers.
For each predictor $s$, we compute  
\begin{itemize}
\item its coverage $\hat{\eta}_s$ over the test set $\mathcal{M}_{\mathrm{test}}$, i.e, the percentage of test data points that are contained in the respective prediction sets:
\begin{align*}
	\hat{\eta}_s = \frac{1}{n_{\mathrm{test}}} \sum_{(x,y)\in\mathcal{M}_{\mathrm{test}}}\mathbbm{1} \{y\in \mathcal{Y}_{s}(x)\},
\end{align*}
where $n_{\mathrm{test}}$ is the number of data points in $\mathcal{M}_{\mathrm{test}}$, and
\item its conservatism $\hat{c}_s$ over the test set $\mathcal{M}_{\mathrm{test}}$, which is evaluated using the average volume of the prediction sets for $\ny\leq5$:
\begin{align}\label{eq:consTest}
	\hat{c}_s = \frac{1}{n_{\mathrm{test}}}\sum_{(x,y)\in\mathcal{M}_{\mathrm{test}}} V(\mathcal{Y}_{s}(x)),
\end{align}  
where $V(\mathcal{Y}_{s}(x))$ computes the volume of the set $\mathcal{Y}_{s}(x)$. 
Since the computation of the exact volume becomes computationally expensive for high-dimensional zonotopes, we evaluate the conservatism of a predictor for $\ny>5$ with
\begin{align*}
	\hat{c}_s = \frac{1}{n_{\mathrm{test}}}\sum_{(x,y)\in\mathcal{M}_{\mathrm{test}}}\sum_{i=0}^{\lceil \ny/3\rceil-1} V_i(\mathcal{Y}_{s}(x)),
\end{align*}
where $V_i(\mathcal{Y}_{s}(x))$ computes the volume of the set $\mathcal{Y}_{s}(x)$ projected onto the dimensions $\{3i+1,3i+2,3i+3\}$.
\end{itemize}
For each data set and predictor, we report the average coverage and conservatism across 30 trained networks, along with their 95\% bootstrap confidence intervals, as shown in \cref{fig:uq}. Example prediction sets for two test data points are illustrated in \cref{fig:uq_sets}.

\ifthenelse{\boolean{showfigure}}{
\begin{figure}[t!]
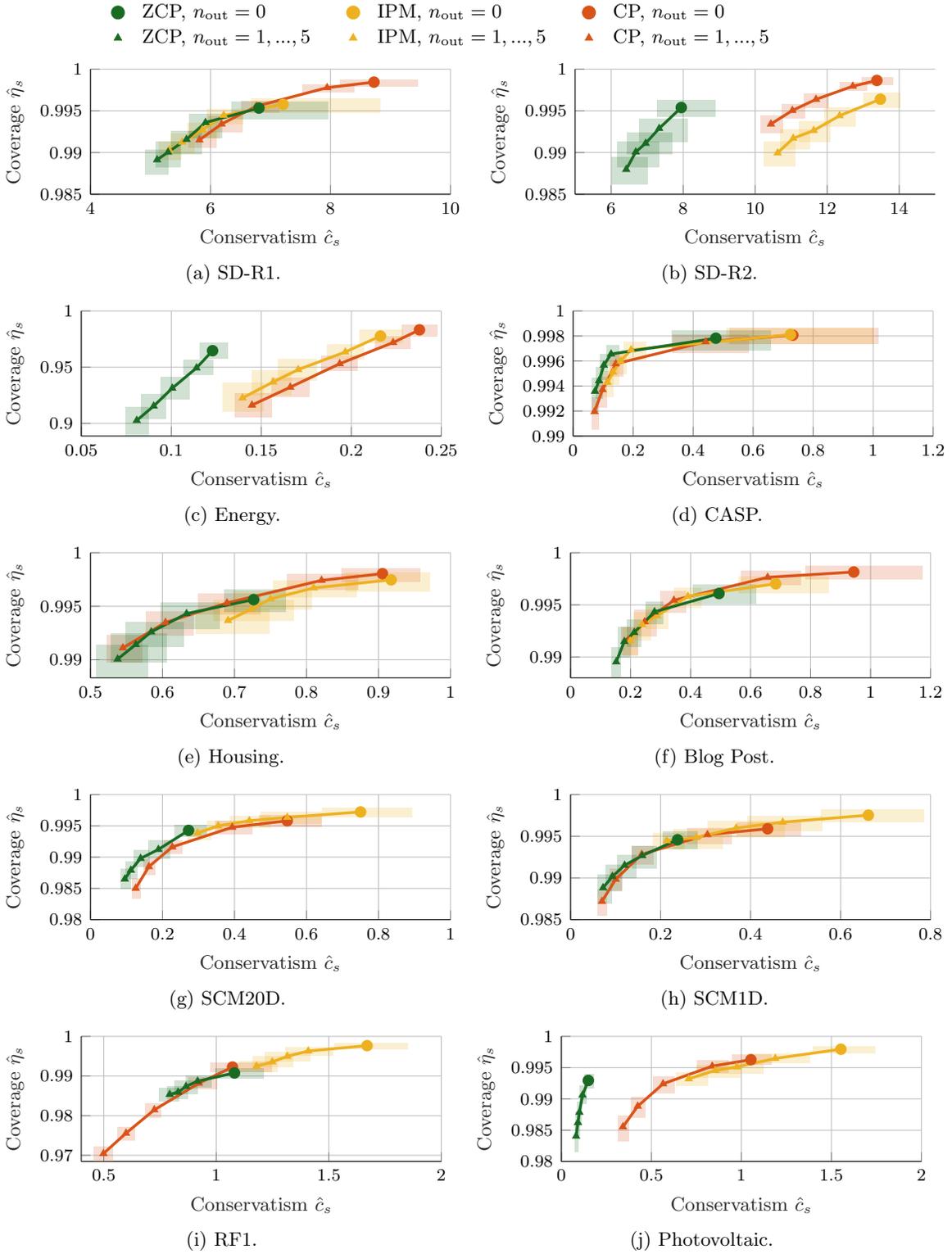

	\centering        
	\input{figures/legendUQ.tikz}
	\vspace{-1em}
	
	\subfloat[SD-R1.\label{fig:uq_sdr1}]{
		\input{figures/tikzUQ_regArtificial}
	}    
	\subfloat[SD-R2.\label{fig:uq_sdr2}]{
		\input{figures/tikzUQ_regArtificialSad}
	}
	
	\vspace{-0.2cm}
	\subfloat[Energy.\label{fig:uq_energy}]{
		\input{figures/tikzUQ_regEnergy}
	}
	\subfloat[CASP.\label{fig:uq_casp}]{
		\input{figures/tikzUQ_regCASP}
	}
	
	\vspace{-0.2cm}
	\subfloat[Housing.\label{fig:uq_housing}]{
		\input{figures/tikzUQ_regHousing}
	}
	\subfloat[Blog Post.\label{fig:uq_blogPost}]{
		\input{figures/tikzUQ_regBlog}
	}
	
	\vspace{-0.2cm}
	\subfloat[SCM20D. \label{fig:uq_SCM20D}]{
		\input{figures/tikzUQ_regSCM20D}
	}
	\subfloat[SCM1D.\label{fig:uq_SCM1D}]{
		\input{figures/tikzUQ_regSCM1D}
	}
	
	\vspace{-0.2cm}
	\subfloat[RF1. \label{fig:uq_rf1}]{
		\input{figures/tikzUQ_regRF1}
	}
	\subfloat[Photovoltaic.\label{fig:uq_Photovoltaic}]{
		\input{figures/tikzUQ_regPhotovoltaic}
	}
	
	\vspace{-0.2cm}
	
	\caption{Trade-off between coverage and conservatism for different regression tasks. 
	}
	\label{fig:uq}
	\vspace{-0.3cm}
	\end{figure}}

	\ifthenelse{\boolean{showfigure}}{
\begin{figure}[t!]
	\centering  
	\vspace{-2em}
	\input{figures/legendUQSets.tikz}
	\vspace{-1.1em}
	
	\subfloat[SD-R1.\label{fig:uqSets_sdr1}]{  
		~~\input{figures/tikzUQSets_regArtificial_train1_i2}        ~~\input{figures/tikzUQSets_regArtificial_train1_i5}
	}~\quad    
	\subfloat[SD-R2.\label{fig:uqSets_sdr2}]{
		~~\input{figures/tikzUQSets_regArtificialSad_train1_i1}        ~~\input{figures/tikzUQSets_regArtificialSad_train1_i2}
	}
	
	\vspace{-0.2cm}
	\subfloat[Energy.\label{fig:uqSets_energy}]{
		\input{figures/tikzUQSets_regEnergy_train1_i1}   \hspace{-0.4cm}     \input{figures/tikzUQSets_regEnergy_train1_i7}
	}
	\subfloat[CASP.\label{fig:uqSets_casp}]{
		\hspace{-0.2cm} \input{figures/tikzUQSets_regCASP_train1_i1}      \hspace{-0.4cm}  \input{figures/tikzUQSets_regCASP_train1_i3}
	}
	
	\vspace{-0.2cm}
	\subfloat[Housing.\label{fig:uqSets_housing}]{
		\input{figures/tikzUQSets_regHousing_train1_i1}        \input{figures/tikzUQSets_regHousing_train1_i5}
	}\quad
	\subfloat[BlogPost.\label{fig:uqSets_blog}]{
		\input{figures/tikzUQSets_regBlog_train1_i1}        \input{figures/tikzUQSets_regBlog_train1_i2}
	}
	
	\vspace{-0.2cm}
	\subfloat[SCM20D. \label{fig:uqSets_SCM20d}]{
		~\input{figures/tikzUQSets_regSCM20D_train1_i1}        ~\input{figures/tikzUQSets_regSCM20D_train1_i9}
	}~
	\subfloat[SCM1D.\label{fig:uqSets_SCM1d}]{
		\input{figures/tikzUQSets_regSCM1D_train1_i1}        \input{figures/tikzUQSets_regSCM1D_train1_i5}
	}
	
	\vspace{-0.2cm}
	\subfloat[RF1.\label{fig:uqSets_rf1}]{
		\input{figures/tikzUQSets_regRF1_train1_i1}    \hspace{-0.4cm}     \input{figures/tikzUQSets_regRF1_train1_i2}
	}~    \subfloat[Photovoltaic.\label{fig:uqSets_photo}]{
		\input{figures/tikzUQSets_regPhotovoltaic_train1_i1}     \hspace{-0.4cm}  \input{figures/tikzUQSets_regPhotovoltaic_train1_i2}
	}
	\vspace{-0.2cm}
	
	\caption{Prediction sets for different regression tasks projected onto the $y_1-y_2$ plane for example data points from the test set. The true outputs are denoted by gray crosses. For the synthetic data sets SD-R1 and SD-R2, we included 100 outputs, which would have been possible for the given input using different uncertainty realizations.}
	\label{fig:uq_sets}
	\vspace{-1em}
	\end{figure}}
	
	Overall, ZCPs produce smaller prediction sets compared to IPMs and CPs. This is largely due to the greater flexibility of zonotopes, which can represent not only axis-aligned intervals but also more complex geometries in the output space.
	ZCPs offer the greatest advantage over CPs and IPMs when the output variables are correlated, as multi-dimensional intervals cannot adequately capture such dependencies. 
	For instance, for the \emph{SD-R2}, \emph{Energy}, and \emph{Photovoltaic} data sets, ZCPs produce significantly smaller prediction sets compared to CPs and IPMs, as illustrated in \cref{fig:uq_sdr2}, \cref{fig:uq_energy}, and \cref{fig:uq_Photovoltaic}. 
	This indicates that, in these cases, ZCPs are able to effectively capture and leverage output correlations, leading to less conservative uncertainty quantification.
	In contrast, for the \emph{SD-R1} and \emph{BlogPost} data sets, the prediction sets produced by ZCPs closely resemble axis-aligned intervals (see \cref{fig:uqSets_sdr1,fig:uqSets_blog}), suggesting that the selected uncertainties fail to capture the inherent correlations. Consequently, as shown in \cref{fig:uq_sdr1} and \cref{fig:uq_blogPost}, ZCPs offer little reduction in conservatism for these data sets.
	
	The increased modeling flexibility of ZCPs comes at the cost of reduced coverage guarantees on unseen data compared to CPs due to a higher risk of overfitting to the calibration set.
	This can also be observed in Tab.~\ref{tab:coverage}, which reports the smallest $\epsilon$ values satisfying~\cref{eq:eta} for a given confidence level. Since we apply CP dimension-wise---identifying one parameter and discarding $\nout$ outliers for each dimension---we set $\ntheta=\ny$ and allow up to $\nout \ny$ outliers in \cref{eq:eta}. 
	Despite this increased allowance for outliers, CPs achieve stronger coverage guarantees, as ZCPs and IPMs typically identify a larger number of parameters ($\ntheta> \ny$).
	The particularly low coverage observed for the \emph{Energy} data set is attributed to its limited calibration set size.

	\begin{table}[t]
\centering
\begin{tabular}{c c c c c c c}
	\toprule
	\textbf{Task} &\multicolumn{2}{c}{$n_{\mathrm{out}}=0$} & \multicolumn{2}{c}{$n_{\mathrm{out}}=1$} & \multicolumn{2}{c}{$n_{\mathrm{out}}=5$}\\
	& CP & IPM/ZCP & CP & IPM/ZCP & CP & IPM/ZCP \\ 
	\cmidrule(r){1-1} \cmidrule(r){2-3} \cmidrule(r){4-5} \cmidrule(r){6-7}
	{Energy} & 0.0496 & 0.2508 & 0.1048 & 0.3291 & 0.2622 & 0.4928 \\ 
    {SCM1d} & 0.0068 & 0.0228 & 0.0183 & 0.0309 & 0.0481 & 0.0495 \\ 
    {SCM20d} & 0.0074 & 0.0249 & 0.0200 & 0.0337 & 0.0524 & 0.0540 \\ 
    {RF1} & 0.0130 & 0.0299 & 0.0421 & 0.0397 & 0.1088 & 0.0619 \\ 
    {Photovoltaic} & 0.0077 & 0.0581 & 0.0207 & 0.0729 & 0.0544 & 0.1040 \\ 
    {Others} & 0.0039 & 0.0201 & 0.0083 & 0.0274 & 0.0216 & 0.0447 \\ 
	\bottomrule
\end{tabular}    
\caption{Smallest values of $\epsilon$ for which the coverage guarantee $\mathbb{P}\{\eta(\mathcal{M}_{\mathrm{cal}}) \geq 1-\epsilon \} \geq 0.9$ is satisfied, across different predictors and prediction tasks. The guarantees for the regression tasks \emph{SD-R1}, \emph{SD-R2}, \emph{CASP}, \emph{Housing}, and \emph{BlogPost} are identical and given in the last row.}
\label{tab:coverage}
\end{table}

\subsection{Classification Tasks}
For classification tasks, where a set of classes is returned if the uncertainty is high, we compare ZCPs with
\begin{itemize}
\item IPMs that are identified using the presented ZCP framework for classification tasks but represent the uncertainties and prediction sets with multi-dimensional intervals, and 
\item CPs that use the classification score function in \cref{eq:cp_addClass_score}. 
\end{itemize}
We identify for each neural network ZCPs, IPMs, and CPs using the calibration data and considering $n_{\mathrm{out}}=0,...,5$ outliers.
Each predictor $s$ is evaluated using
\begin{itemize}
\item its coverage $\hat{\eta}_s$ over the test set $\mathcal{M}_{\mathrm{test}}$, i.e, the percentage of test data points where the corresponding prediction set contains the correct class:
\begin{align*}
	\hat{\eta}_s = \frac{1}{n_{\mathrm{test}}} \sum_{(x,y)\in\mathcal{M}_{\mathrm{test}}}\mathbbm{1} \Bigl\{\mathrm{classes}(y)\subseteq \mathrm{classes}\bigl(\mathcal{Y}_{s}(x)\bigr)\Bigr\},
\end{align*}
where $n_{\mathrm{test}}$ is the number of data points in $\mathcal{M}_{\mathrm{test}}$, and
\item its conservatism $\hat{c}_s$ over the test set $\mathcal{M}_{\mathrm{test}}$, which is evaluated using the average number of predicted classes:
\begin{align*}
	\hat{c}_s = \frac{1}{n_{\mathrm{test}}}\sum_{(x,y)\in\mathcal{M}_{\mathrm{test}}} V\Bigl(\mathrm{classes}\bigl(\mathcal{Y}_{s}(x)\bigr)\Bigr),
\end{align*}  
where $V\Bigl(\mathrm{classes}\bigl(\mathcal{Y}_{s}(x)\bigr)\Bigr)$ returns the number of elements in $\mathrm{classes}\bigl(\mathcal{Y}_{s}(x)\bigr)$. 
\end{itemize}
For each data set and predictor, we plot the coverage over the conservatism averaged over all 30 networks as well as their 95\% bootstrap confidence intervals in \cref{fig:uqClass}.
Similar to the regression setting, ZCPs produce less conservative prediction sets than both IPMs and CPs in the classification setting. 
This benefit is particularly evident for the \emph{MNIST} and \emph{Covertype} data sets, where ZCPs predict substantially fewer classes while allowing the same number of outliers. 
However, this reduction in conservatism comes at the cost of a slight decrease in coverage.

Furthermore, \cref{fig:uqClass_sets} shows representative prediction sets for IPMs and ZCPs on individual test samples. Since conformal prediction directly yields a set of classes without score-based uncertainty regions, the results of CPs are omitted from this figure.
For each input $x$ in \cref{fig:uqClass_sets}, IPMs and ZCPs generate prediction sets $\mathcal{Y}(x)$, which intersect the gray area, indicating there exists a score vector $y\in\mathcal{Y}(x)$ consistent with correct classification. 
For the synthetic data sets \emph{SD-C1} and \emph{SD-C2}, both IPMs and ZCPs often produce compact prediction sets fully contained within the gray area (see left columns of \cref{fig:uqSets_sdc1,fig:uqSets_sdc2}), reflecting high classification confidence. 
Furthermore, in the right columns of \cref{fig:uqSets_sdc1,fig:uqSets_covtype}, we see that although the deterministic prediction model $f(x)$ alone would misclassify the sample (as indicated by the black asterix lying outside the gray region), both IPMs and ZCPs correctly account for the uncertainty by generating sets that intersect the gray region---signaling the potential for correct classification.
The advantage of zonotopes becomes apparent in the right column of \cref{fig:uqSets_mnist}: while IPMs generate sets that extend into the white region---suggesting ambiguity---the ZCP with $n_{\mathrm{out}}=5$ results in a prediction set that remains strictly within the gray region, thus, excluding the incorrect class.

\ifthenelse{\boolean{showfigure}}{
\begin{figure}[p]
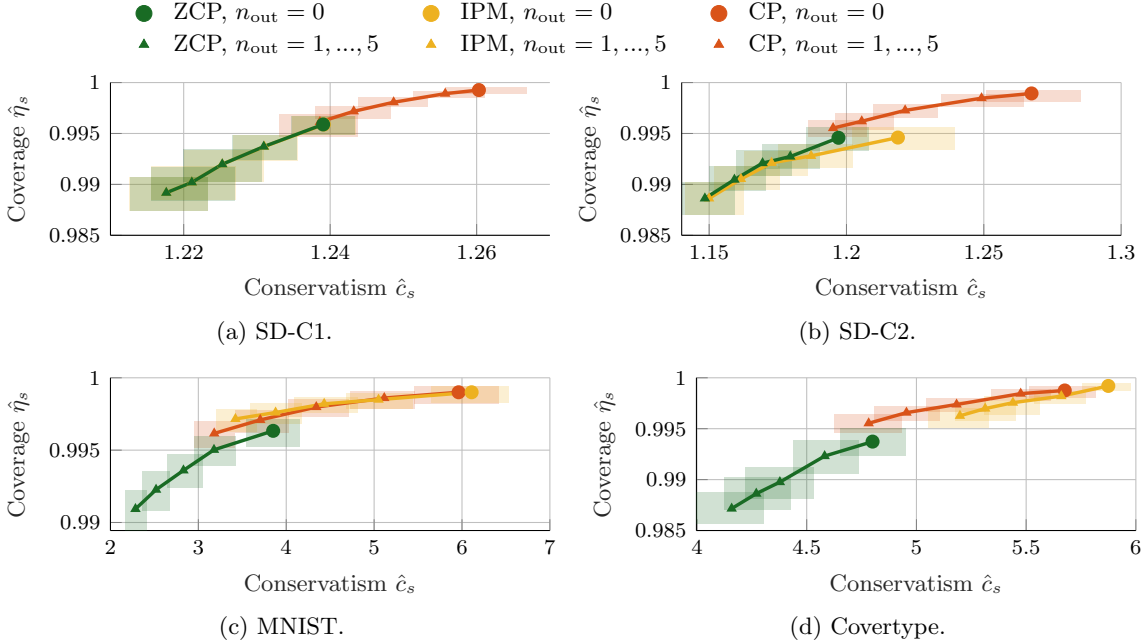

	\centering        
	\vspace{-2em}
	\input{figures/legendUQ.tikz}
	\vspace{-1em}
	
	\subfloat[SD-C1.\label{fig:uq_sdc1}]{
		\input{figures/tikzUQ_classArtificial1}
	}    
	\subfloat[SD-C2.\label{fig:uq_sdc2}]{
		\input{figures/tikzUQ_classArtificial2}
	}
	
	\vspace{-0.2cm}
	\subfloat[MNIST.\label{fig:uq_mnist}]{
		\input{figures/tikzUQ_classMNIST}
	}
	\subfloat[Covertype.\label{fig:uq_covtype}]{
		\input{figures/tikzUQ_classCovtype}
	}

	\vspace{-0.2cm}
	
	\caption{Trade-off between coverage and conservatism for different classification tasks. 
	}
	\label{fig:uqClass}
\end{figure}

\begin{figure}[p]
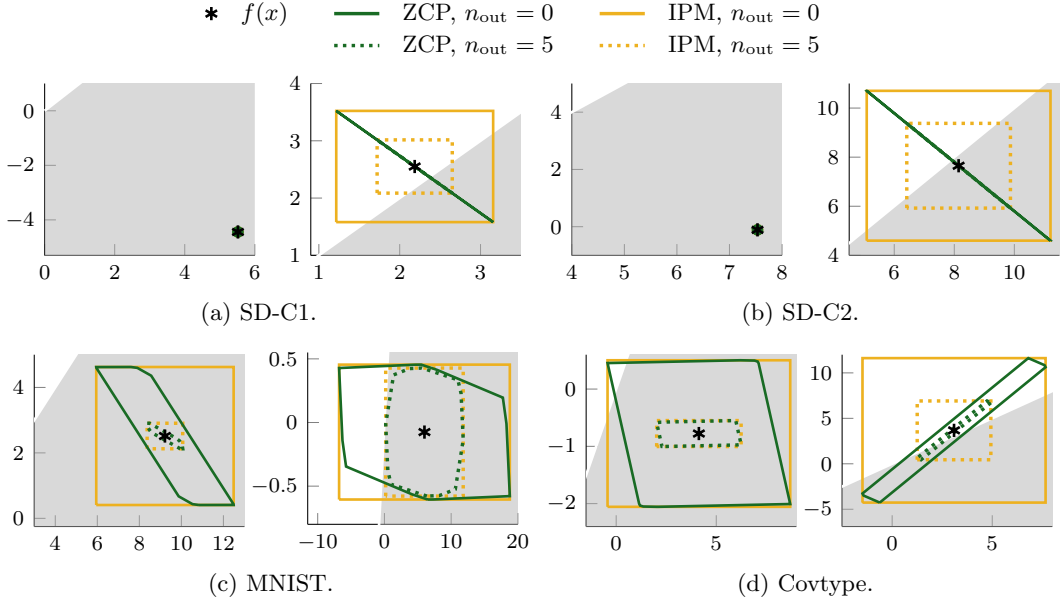

	\centering  
	\input{figures/legendUQSets_class.tikz}
	\vspace{-1em}
	
	\subfloat[SD-C1.\label{fig:uqSets_sdc1}]{  
		\input{figures/tikzUQSets_classArtificial1_train5_i2}    \input{figures/tikzUQSets_classArtificial1_train5_i1}
	}  
	\subfloat[SD-C2.\label{fig:uqSets_sdc2}]{
		\input{figures/tikzUQSets_classArtificial2_train2_i1}      \input{figures/tikzUQSets_classArtificial2_train2_i8}
	}
	
	\vspace{-0.2cm}
	\subfloat[MNIST.\label{fig:uqSets_mnist}]{
		\input{figures/tikzUQSets_classMNIST_train5_i4}   \hspace{-0.3cm}     \input{figures/tikzUQSets_classMNIST_train5_i5}
	}
	\subfloat[Covtype.\label{fig:uqSets_covtype}]{
		\hspace{-0.4cm} \input{figures/tikzUQSets_classCovtype_train1_i7}    \hspace{-0.3cm}  \input{figures/tikzUQSets_classCovtype_train1_i3}
	}
	\vspace{-0.2cm}
	
	\caption{Prediction sets for classification tasks for example data points from the test set. The $x$-axis represents the output score for the true class, while the $y$-axis corresponds to the output for an arbitrarily chosen incorrect class. The gray shaded region marks the domain where the true class has a higher score than the incorrect class.}
	\label{fig:uqClass_sets}
	\vspace{-1em}
	\end{figure}}
	
	\section{Conclusions} \label{sec:conclusions}
	
	We have introduced zono-conformal predictors---a novel framework for uncertainty quantification that builds methodologically on interval predictor models, while aligning with conformal prediction in its practical goal of equipping arbitrary predictors with reliable uncertainty sets.
	Unlike conventional approaches that rely on axis-aligned intervals or separate data sets for the uncertainty modeling and calibration steps, we construct prediction zonotopes by identifying uncertainties directly within the predictor architecture. 
	This allows the uncertainty modeling and calibration to be unified into a single, data-efficient linear program, while preserving valid coverage guarantees for unseen data.
	We also extended ZCPs to classification tasks and developed strategies for outlier detection that further reduce conservatism by identifying and removing anomalous data points during calibration. 
	Our evaluation on different data sets demonstrates that ZCPs consistently yield smaller prediction sets than CPs and IPMs, particularly in multi-output settings with correlated outputs---scenarios where conventional interval-based methods struggle to capture dependencies effectively.
	Beyond test-time performance, ZCPs offer important practical benefits in application domains where calibration data are reused, such as in safety-critical learning-based systems. 
	Since ZCPs maintain the same coverage over calibration data as IPMs and CPs but generate significantly smaller prediction sets, they allow for tighter safety margins without compromising safety. 	
	Nevertheless, ZCPs also come with certain limitations: 
    \begin{itemize}
        \item ZCPs are calibrated by solving an optimization problem, resulting in increased computational cost compared to the calibration of standard CPs.
        \item The guaranteed coverage on unseen data degrades as the number of identified uncertainties increases due to potential overfitting.
        \item Zonotopes are restricted to convex and centrally symmetric shapes. Although our framework naturally generalizes to other set representations, a linear-programming-based construction requires that both a proxy for set volume and the containment condition $y \in \mathcal{Y}(x)$ remain linear in the parameters $\theta$.
    \end{itemize}
	Future work will focus on mitigating these limitations. 
    In particular, we plan to develop training strategies for the base predictor that promote a small set of uncertainties at user-defined locations, thereby reducing computational overhead in the calibration step and improving coverage guarantees. 
    In parallel, exploring alternative set representations and parameterizations---capable of expressing multi-modal or non-convex uncertainties beyond a single zonotope---offers a promising avenue for enhancing the applicability of the ZCP framework.     
	Overall, zono-conformal prediction provides a flexible, scalable, and principled framework for reliable uncertainty quantification, offering a strong foundation for safer and more effective deployment of machine learning models in real-world systems.
	\acks{
The authors would like to thank the anonymous reviewers and the action editor for their constructive comments. The authors also want to thank Jan Jakub Kamiński for his valuable feedback on the manuscript. 
This work was funded by the German Research Foundation---SFB 1608 under grant number 501798263 and grant number 458030766---and supported by a fellowship within the IFI programme of the German Academic Exchange Service.}

\newpage

\appendix
\section{Data Sets} \label{sec:datasets}

For regression tasks, we consider two synthetic and eight real-world data sets, as listed in Tab.~\ref{tab:data}. 
All data points of the real-world data sets are normalized to lie between 0 and 1.
The synthetic and the \emph{Photovoltaic} data sets are described in the following:
\begin{itemize}
\item The synthetic data set \emph{SD-R1} contains data generated by the test function
\begin{align*}
g(x,\unc) = \begin{bmatrix}
	5\sin(x_1)+x_2^2 + x_1 \unc_1\\
	\frac{1}{x_1^2+1}+\cos(x_2) + x_2 \unc_2
\end{bmatrix},
\end{align*}
where $x$ is randomly sampled from the multi-dimensional interval $[-5\cdot\mathbf{1},~5\cdot\mathbf{1}]$ and $u$ is sampled from the zonotope $\langle \mathbf{0},~[0.2\cdot \mathbf{1}~~~0.02\cdot v_1]\rangle$, where $v_1=[-0.209~~ 1.129]^\top$ was randomly sampled from a standard normal distribution.

\item The synthetic data set \emph{SD-R2} contains data generated by the test function \citep{sadeghi2019efficientTraining}
\begin{align*}
g(x,\unc) &=\begin{bmatrix}            3x_1^3+\exp(\cos(10x_2)\cos^2(5x_1))\\            2x_1^2+\exp(\cos(10x_1)\cos^2(5x_2))
\end{bmatrix} \\
&\qquad +\begin{bmatrix}
	\exp(\sin(7.5x_3))+\unc_1 \\ 
	\exp(\sin(7.5x_3^2))+ 1.5\unc_2
\end{bmatrix},
\end{align*} 
where $x$ is randomly sampled from the multi-dimensional interval $[\mathbf{0},~\mathbf{1}]$ and $u$ is sampled from the zonotope $\langle 0.5\cdot\mathbf{1},~[0.5\cdot \mathbf{1}~~~0.05\cdot v_2]\rangle$, where $v_2=[0.747~~ -0.247]^\top$ was randomly sampled from a standard normal distribution.
\item The \emph{Photovoltaic} data set is based on the hourly power generation profile of a residential photovoltaic installation from the SimBench data set \textit{1-LV-rural2–1-sw} \citep{meinecke2020simbench}. The predictor is trained to predict the next four hours based on the past 48 hours. This is a common task in energy management and serves as an example where the output features are highly correlated.
\end{itemize}
For classification tasks, we consider two synthetic and two real-world data sets, as listed in Tab.~\ref{tab:data}. 
The two synthetic data sets are generated as follows, where the test functions are designed to have intersections, i.e., certain inputs $x$ could have been generated by multiple test functions:
\begin{itemize}
\item We generate the synthetic data set \emph{SD-C1} from the test functions
\begin{align*}
\text{class 1: }x_2 &= 3 \sin(x_1) + \unc_1, \\
\text{class 2: }x_2 &= x_1^2 + \unc_1, \\
\text{class 3: }x_2 &= 2x_1-10 + \unc_1,
\end{align*}
where $x_1$ and $\unc_1$ of each class are randomly sampled from the intervals $[-5,~5]$ and $[-2,~2]$, respectively.
\item We generate the synthetic data set \emph{SD-C2} from the test functions
\begin{align*}
\text{class 1: }x_3 &= x_2\sin(x_1)+\unc_1, \\
\text{class 2: }x_3 &= x_1^2+x_2+2\unc_1, \\
\text{class 3: }x_3 &= 2x_1-10+x_1x_2+0.5\unc_1^2, \\
\text{class 4: }x_3 &= 2x_1-16+x_2\unc_1,
\end{align*}
where $x_1$, $x_2$, and $\unc_1$ of each class are randomly sampled from the intervals $[-5,~5]$, $[-5,~5]$, and $[-1,~1]$, respectively.
\end{itemize}

\section{Ablation Studies}\label{sec:ablationStudies}
In this section, we evaluate the impact of different uncertainty placement strategies, cost functions, and outlier detection methods within the zono-conformal prediction framework. To manage computational complexity, we limit the calibration data sets to a maximum of 300 data points.
Unless stated otherwise, ablation results are reported for the regression data sets \emph{SD-R1}, \emph{SD-R2}, \emph{Energy}, and \emph{SCM20D}. However, we observed similar qualitative trends for uncertainty placement and outlier detection across classification data sets as well.
To enable consistent comparisons across different data sets and methods, we introduce the following normalized evaluation metrics:
\begin{itemize}
\item The normalized conservatism of a predictor $s$ with respect to a baseline predictor $\tilde{s}$ over the calibration data can be computed with
\begin{align}\label{eq:consTest}
\hat{c}_{s/\tilde{s}} = \frac{1}{n_{\mathrm{m}}}\sum_{{(x,y)\in\mathcal{M}_{\mathrm{cal}}}} \frac{V(\mathcal{Y}_{s}(x))}{V(\mathcal{Y}_{\tilde{s}}(x))},
\end{align}  
where $V(\mathcal{Y}_{s}(x))$ denotes the volume of the prediction set $\mathcal{Y}_{s}(x)$. 
\item The normalized computation time of a predictor $s$ with respect to a baseline predictor $\tilde{s}$ is defined as
\begin{align*}
T_{s/\tilde{s}} = \frac{T_s(\mathcal{M}_{\mathrm{cal}})}{T_{\tilde{s}}(\mathcal{M}_{\mathrm{cal}})},
\end{align*}
where $T_s(\mathcal{M}_{\mathrm{cal}})$ is the total time required for constructing predictor $s$ from the calibration data set $\mathcal{M}_{\mathrm{cal}}$.
\end{itemize}
In all figures, methods highlighted in \textcolor{mycolor3}{green} represent those we consider most practical based on the trade-off between performance and efficiency. These are the default choices adopted throughout the remainder of this work.
Methods shown in \textcolor{mycolor6}{purple} correspond to naïve baselines or existing approaches from the literature, while those in shades of \textcolor{mycolor5}{blue} denote other novel strategies introduced in this paper.

\subsection{Uncertainty Placement Strategies}\label{subsec:ablationUP}

We compare several strategies for selecting a subset of uncertainties $\unc \in \mathbb{R}^{\nunc}$, $\nunc=\round{p_{\mathrm{p}}n_{\mathrm{p}}}+\ny$, from the candidate uncertainties $\tilde{\unc}=[\unc_{\mathrm{p}}^\top~\unc_{\mathrm{y}}^\top]^\top\in \mathbb{R}^{n_{\mathrm{p}}+\ny}$, with the parameter $p_{\mathrm{p}}\in[0,1]$.
The parametric uncertainties $u_{\mathrm{p}}$ are added to all biases of the hidden layers of the neural network.
Unless stated otherwise, we use the identity matrix as the generator template $G_{\mathrm{\unc}}$, which yields $\ntheta=\nunc$ parameters to be identified.
We consider the following strategies:
\begin{itemize}
\item \emph{ORand}: We select all output uncertainties $\unc_{\mathrm{y}}$ and a random selection of $\round{p_{\mathrm{p}}n_{\mathrm{p}}}$ parametric uncertainties, as proposed in \cref{subsec:uncertaintyPlacement}.
\item \emph{ORand*}: We select the uncertainties as in \emph{ORand} (see \cref{subsec:uncertaintyPlacement}), but create the generator template $G_{\mathrm{\unc}}$ from a horizontal concatenation of the identity matrix and $\nunc$ random generators, i.e., $\ntheta=2\nunc$.
\item \emph{QR}: We select $\nunc$ uncertainties using the following deterministic QR-factorization strategy, which aims to select uncertainties whose influence is as different as possible up to some linear scaling factor.
\begin{enumerate}
\item 
For each candidate uncertainty $\tilde{\unc}_{(i)}$, we compute the partial derivative $\frac{\partial \tilde{f}(x,\tilde{\unc})}{\partial \tilde{\unc}_{(i)}}|_{\tilde{\unc}=\mathbf{0}}$, where the partial derivative is evaluated at $\tilde{\unc}=\mathbf{0}$ as we assume the uncertainties are small and zero-centered.
We construct the matrix $V$ by concatenating the derivatives for the inputs $x^{(m)}$, $m=1,\dots,\nm$, and uncertainties $\tilde{\unc}_{(i)}$, where the $i$-th column of $V$ is given by
\begin{align*}
	V_{(\cdot,i)} = \mathrm{vert}_{ m\in\{1,\dots,\nm\}}\left(\frac{\partial f(x^{(m)},\tilde{\unc})}{\partial \tilde{\unc}_{(i)}}|_{\tilde{\unc}=\mathbf{0}}\right)
\end{align*}    
and describes the general influence of uncertainty $\tilde{\unc}_{(i)}$ on the output.
\item We apply the QR-factorization algorithm with column pivoting \citep{businger1965,engler1997} on the matrix $V$ to greedily select columns $V_{(\cdot,i)}$ that are as linearly independent from the previous ones as possible.
\item 
We select all output uncertainties by default plus the uncertainties corresponding to the first $\nunc-\ny$ columns of $V$ (excluding any already-selected output uncertainties).
\end{enumerate}
Since this strategy relies on the calibration data set, the resulting coverage guarantees are weaker than those of data-independent strategies. As an alternative, a separate data set could be employed specifically for uncertainty placement.
\item \emph{Rand}: We randomly select $\round{p_{\mathrm{p}}n_{\mathrm{p}}}+\ny$ uncertainties from $\tilde{u}$.
\end{itemize}

The normalized conservatism and computation time are visualized in \cref{fig:up} for varying identification ratios $p_{\mathrm{p}}$.
The more uncertainties we identify by increasing $p_{\mathrm{p}}$, the lower is the conservatism of the prediction sets but the higher is the computation time.
Strategies that include all output uncertainties by default---\emph{ORand}, \emph{ORand*}, and \emph{QR}---consistently yield less conservative prediction sets compared to the fully random strategy \emph{Rand}.
Interestingly, the simple \emph{ORand} approach leads to even smaller prediction sets than the more complex \emph{QR} method.
Adding extra random generators in the generator template $G_{\mathrm{\unc}}$, as done in \emph{ORand*}, slightly reduces conservatism but increases computational time.
Overall, the \emph{ORand} strategy offers the best trade-off between low conservatism and computational efficiency. Since selecting too many uncertainties can negatively impact the coverage guarantees (see \cref{sec:coverage}), we use $p_{\mathrm{p}}=0.1$ for all other experiments.

\ifthenelse{\boolean{showfigure}}{
\begin{figure}[t]
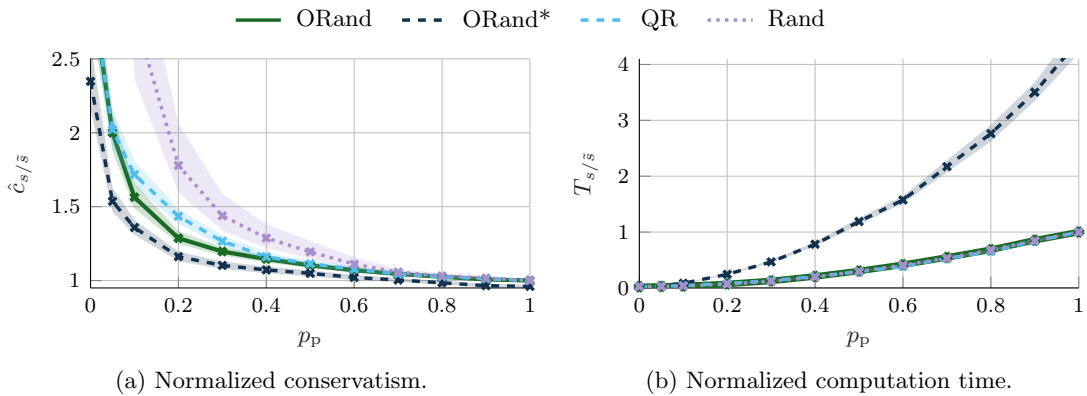

\centering        
\input{figures/legendUP.tikz}
\vspace{-1em}

\subfloat[Normalized conservatism.\label{fig:upFval}]{
	\centering        
	\input{figures/tikzUP_cons}
	
}
\subfloat[Normalized computation time.\label{fig:upT}]{
	\centering        
	\input{figures/tikzUP_time}
	
}

\caption{Comparison of uncertainty placement strategies. The values are normalized by the results of the predictor $\tilde{s}$, identified using \emph{ORand} with $p_{\mathrm{p}}=1$, and averaged over all networks and data sets. The shaded regions represent 95\% bootstrap confidence intervals.}
\label{fig:up}
\vspace{-0.4cm}
\end{figure}}

\subsection{Cost Functions for Regression} \label{subsec:costRegression}

In regression tasks, our goal is to identify uncertainties that both satisfy the conformance constraints and minimize the volume of the resulting prediction sets. 
The volume of a zonotope $\mathcal{Z}=\langle c, G\rangle$, with $G\in \mathbb{R}^{n \times {\nu}}$, is given by~\citep[Cor.~3.4]{gover2010parallelotope}
\begin{align} \label{eq:vol}
\text{volume}(\mathcal{Z}) = 2^{n} 
\sum_{1\leq j_1 <\dots<j_{n}\leq\nu} |\mathrm{det}(G_{j_1,\dots,j_{n}})|,
\end{align}
where the summation considers all possible choices of $j_1,\dots,j_n$ that satisfy $1 \leq j_1 < \dots < j_{n} \leq \nu$, and
$G_{j_1,\dots,j_{n}}$ is constructed from columns $j_1,\dots,j_{n}$  of 
$G$ for all possible choices of $j_1,\dots,j_{n}$ that satisfy $1 \leq j_1 < \dots < j_{n} \leq \nu$.
However, the volume of a zono-conformal prediction set is non-convex in the scaling factors $\alpha$, making direct volume minimization computationally intractable. 
Instead, we evaluate alternative cost functions that encourage small prediction sets and remain linear in $\alpha$:
\begin{itemize}
\item \emph{Interval}: We minimize the interval norm of the prediction sets, which is the standard cost function in reachset-conformant identification \citep{Liu2023conf,luetzow2024generator}, i.e.,
\begin{align}\label{eq:costInterval}
\mathrm{size}(\YZCP(x)) = \|\YZCP(x)\|_I.
\end{align}
\item \emph{Rotated Intervals (RI)}: The interval norm only motivates to minimize the extensions of the prediction sets along the unit vectors. To also reward small extensions along other directions, we propose to minimize the summed interval norms of the prediction sets that are randomly rotated.
This leads to the cost function in \cref{eq:costRotInterval}.
In the following experiments, we evaluate $n_{\mathrm{r}}=5$, $n_{\mathrm{r}}=10$, and $n_{\mathrm{r}}=20$.
\item \emph{Generator Lengths}: Alternatively, we can minimize the summed length of all generators of $\YZCP(x)=\langle c, [g_1~g_2~\cdots~g_{\ntheta}]\rangle$:
\begin{align}\label{eq:costGenLen}
\mathrm{size}(\YZCP(x)) = \sum_{i=1}^{\ntheta} \|g_i\|_2.
\end{align}
\end{itemize}

The normalized conservatism for different data sets is visualized in \cref{fig:cost}, and the unnormalized computation times are reported in Tab.~\ref{tab:cost}.
Among all methods, the \emph{RI} cost function consistently yields the smallest prediction sets, reducing the volume by up to a factor of 0.66 compared to the standard \emph{Interval} cost function. While increasing the number of rotations $n_{\mathrm{r}}$ slightly improves performance, it also increases computation time due to the cost of generating random orthogonal matrices $R_i$. 
Nevertheless, since this overhead does not depend on the calibration set size $\nm$, the \emph{RI} approach scales well with larger data sets.
In fact, when using the full calibration set (i.e., without limiting 
$\nm\leq300$), the average computation time for \emph{RI} with $n_{\mathrm{r}}=10$ was lower than for the \emph{Interval} cost on all data sets except the small \emph{Energy} data set.
Given the marginal improvements beyond $n_{\mathrm{r}}=10$, we adopt this value for all other experiments.

\ifthenelse{\boolean{showfigure}}{
\begin{figure}[t]
\centering        

\input{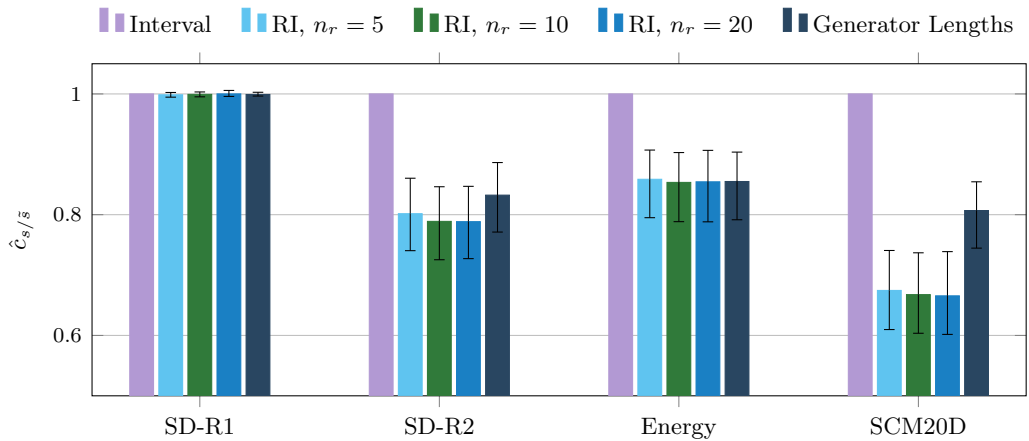}
\caption{Comparison of cost functions for regression. The values are normalized by the results of the predictor $\tilde{s}$, identified using the \emph{Interval} cost function, and averaged over all networks. The error bars denote the 95\% bootstrap confidence intervals.}
\label{fig:cost}
\vspace{-0.4cm}
\end{figure}}

\begin{table}[t]
\centering
\begin{tabular}{c c c c c}
\toprule
\textbf{Cost Function} & \emph{SD-R1} & \emph{SD-R2} & \emph{Energy} & \emph{SCM20D} \\ 
\midrule
\emph{Interval} & 0.52 & 0.55 & 0.11 & 0.77 \\
\emph{RI, $n_r=5$} & 0.55 & 0.66 & 0.11 & 0.83 \\
\emph{RI, $n_r=10$} & 0.61 & 0.66 & 0.13 & 0.81 \\
\emph{RI, $n_r=20$} & 0.74 & 0.77 & 0.16 & 1.03 \\
\emph{Generator Lengths} & 0.47 & 0.51 & 0.09 & 0.68 \\
\bottomrule
\end{tabular}    
\caption{Computation time in seconds using different cost functions.}
\label{tab:cost}
\end{table}

\subsection{Cost Functions for Classification}\label{subsec:costClass}
In classification tasks, the objective is to minimize the number of classes predicted for each input. As smaller prediction sets reduce the likelihood of including multiple classes, we can apply the same cost functions as in the regression setting (see \cref{subsec:costRegression}).
Additionally, we evaluate two cost functions specifically designed to penalize incorrect class scores. 
Specifically, we compare the following cost functions, which can all be formulated linearly in the scaling factors $\alpha$:
\begin{itemize}
\item \emph{Interval}: We minimize the interval norm as in \cref{eq:costInterval}.
\item \emph{Rotated Intervals (RI)}: We minimize the summed interval norms of randomly rotated prediction sets as in \cref{eq:costRotInterval}, where we use  $n_{\mathrm{r}}=10$ rotations.
\item \emph{Generator Lengths}: We minimize the summed generator lengths as in \cref{eq:costGenLen}.
\item \emph{Score}: Alternatively, we can minimize the maximum score of incorrect classes using the cost function
\begin{align}\label{eq:class_cost1}
\mathrm{size}(\YZCP(x)) = \sum_{i\not\in \mathrm{classes}(y)} \max_{z\in \YZCP(x)} z_{(i)},
\end{align}
where $y$ encodes the correct classes for the input $x$.
\item \emph{Score Difference}: To motivate robust correct predictions, we minimize the maximum score difference between incorrect and correct classes, i.e.,
\begin{align}\label{eq:class_cost1}
\mathrm{size}(\YZCP(x),y) = \sum_{i\not\in \mathrm{classes}(y)} \sum_{j\in\mathrm{classes}(y)} d_{ij},
\end{align}
where the score difference $d_{ij}=\max_{z\in \YZCP(x)} z_{(i)}-z_{(j)}$ is the maximum difference between the classification score for class $i$ and the score for class $j$ over the prediction set $\YZCP(x)$.
\end{itemize}
The normalized conservatism across the classification data sets is shown in \cref{fig:costClass}. All evaluated cost functions yield similarly sized prediction sets, with minor variations depending on the data set.
For methodological consistency and simplicity, we adopt the same cost function used in the regression setting---namely, \emph{RI} with $n_r=10$---for all classification experiments.

\ifthenelse{\boolean{showfigure}}{
\begin{figure}[t]
\centering        
   
\input{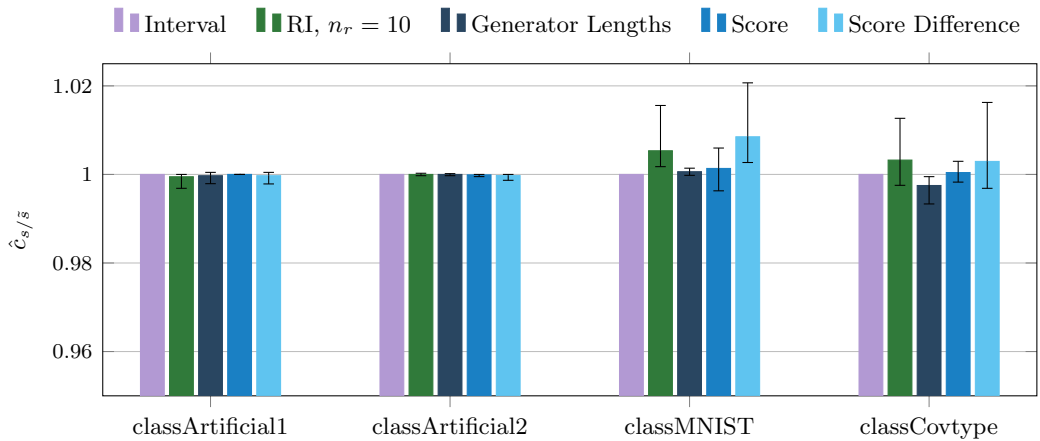}
\caption{Comparison of cost functions for classification. The values are normalized by the results of the predictor $\tilde{s}$, identified using the \emph{Interval} cost function, and averaged over all networks. The error bars denote the 95\% bootstrap confidence intervals.}
\label{fig:costClass}
\vspace{-0.4cm}
\end{figure}}

\subsection{Outlier Detection} \label{subsec:ablationOD}

We compare the following outlier detection methods, each aiming to identify and remove $\nout$ measurements from the calibration data set that most effectively reduce the identification cost:
\begin{itemize}
\item \emph{Search}: the exhaustive search over boundary points as described in \cref{subsubsec:boundary},
\item \emph{SearchG}: the greedy search over boundary points as described in \cref{subsubsec:boundaryGreedy},
\item \emph{MILP}: the mixed-integer linear programs as described in \cref{subsubsec:milp},
\item \emph{RMSE}: a heuristic method that removes the $\nout$ data points with the largest root-mean-square prediction error $\sqrt{e^{(m)\top} e^{(m)}}$, with $e^{(m)}=y^{(m)}-f(x^{(m)})$.
\end{itemize}
\cref{fig:od} shows the normalized conservatism and the normalized computation time for different numbers of removed outliers $\nout$.
The exhaustive search strategy \emph{Search} leads to the optimal identification cost but is computationally feasible only for small $\nout$.
The greedy variant \emph{SearchG} substantially reduces computation time while maintaining near-optimal identification performance.
The \emph{MILP} method consistently reaches optimal identification costs and is more efficient than \emph{Search}, though it leads to slightly higher computation times than \emph{SearchG}.
In contrast, the \emph{RMSE}-based approach is the most computationally efficient but results in noticeably higher identification costs.
Given its favorable trade-off between accuracy and efficiency, we adopt \emph{SearchG} as the default outlier detection method in all other experiments.

\ifthenelse{\boolean{showfigure}}{
\begin{figure}[t]
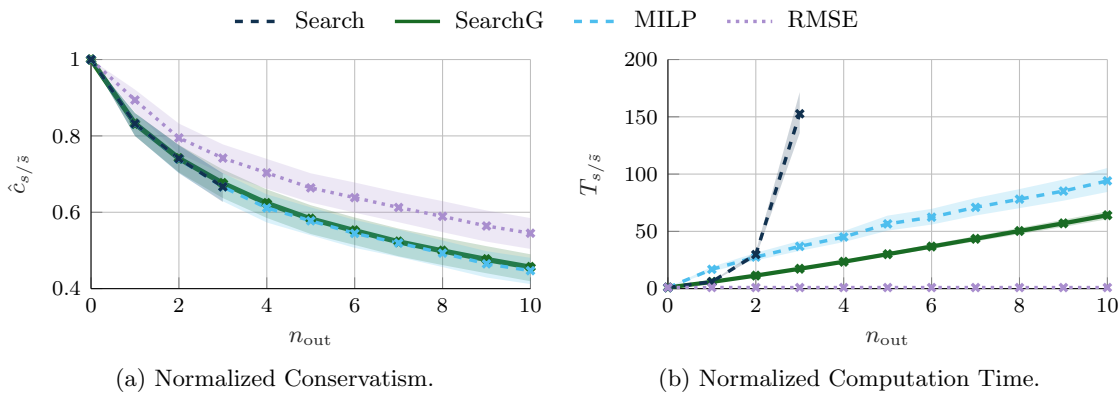

\centering        
\input{figures/legendOD.tikz}
\vspace{-1em}

\subfloat[Normalized Conservatism.\label{fig:odFval}]{
	\input{figures/tikzOD_cons} 
}
\subfloat[Normalized Computation Time.\label{fig:upT}]{
	\input{figures/tikzOD_time}
}

\caption{Comparison of outlier detection methods. The values are normalized by the results of the predictor $\tilde{s}$, which is identified without removing any measurements, i.e., $\nout=0$, and averaged over all networks and data sets. The shaded regions represent 95\% bootstrap confidence intervals.}
\label{fig:od}
\vspace{-0.4cm}
\end{figure}}

\vskip 0.2in
\bibliography{literature}

\end{document}